\pdfoutput=1

\documentclass[11pt]{article}

\usepackage[final]{acl}
\usepackage{times}
\usepackage{latexsym}
\usepackage[T1]{fontenc}
\usepackage[utf8]{inputenc}
\usepackage{microtype}
\usepackage{inconsolata}
\usepackage{graphicx}
\usepackage{color}
\usepackage{xspace}
\usepackage{placeins}
\usepackage{macros}

\title{Quantile Regression with Large Language Models for Price Prediction}

\author{Nikhita Vedula\textsuperscript{\rm 1}$^*$ ~~ Dushyanta Dhyani\textsuperscript{\rm 1}$^*$ ~~ Laleh Jalali\textsuperscript{\rm 1}
\AND
Boris Oreshkin\textsuperscript{\rm 1} ~~ Mohsen Bayati\textsuperscript{\rm 1,2} ~~ Shervin Malmasi\textsuperscript{\rm 1}
\\\\
  \textsuperscript{\rm 1} Amazon.com, Inc. 
  \textsuperscript{\rm 2} Stanford University \\
  \tt \{veduln, dhyanidd, lalehjal, oreshkin, bayatim, malmasi\}@amazon.com \\
}

\usepackage{cleveref}

\definecolor{mydarkgreen}{HTML}{009900}

\begin{document}
\maketitle
\def\thefootnote{*}\footnotetext{~Equal contribution.}
\def\thefootnote{\arabic{footnote}}

\begin{abstract}
Large Language Models (LLMs) have shown promise in structured prediction tasks, including regression, but existing approaches primarily focus on point estimates and lack systematic comparison across different methods.
We investigate probabilistic regression using LLMs for unstructured inputs, addressing challenging text-to-distribution prediction tasks such as price estimation where both nuanced text understanding and uncertainty quantification are critical.
We propose a novel quantile regression approach that enables LLMs to produce full predictive distributions, improving upon traditional point estimates. Through extensive experiments across three diverse price prediction datasets, we demonstrate that a Mistral-7B model fine-tuned with quantile heads significantly outperforms traditional approaches for both point and distributional estimations, as measured by three established metrics each for prediction accuracy and distributional calibration.
Our systematic comparison of LLM approaches, model architectures, training approaches, and data scaling reveals that Mistral-7B consistently outperforms encoder architectures, embedding-based methods, and few-shot learning methods.
Our experiments also reveal the effectiveness of LLM-assisted label correction in achieving human-level accuracy without systematic bias. Our curated datasets are made available\footnote{\scriptsize{\url{https://github.com/vnik18/llm-price-quantile-reg/}}} to support future research.
\end{abstract}

\section{Introduction}
\label{sec:intro}

Large Language Models (LLMs) have demonstrated remarkable capabilities across a wide range of tasks, including unstructured document processing \citep{ZOU2025144572}, instruction-following \citep{ouyang2022training}, and multimodal reasoning and general intelligence benchmarks \citep{openai2023gpt4,bubeck2023sparks}. Going beyond their original purpose of text generation \citep{brown2020language}, they have recently been extended to structured numerical prediction tasks such as time series forecasting \citep{das2024a}.
Recent research has shown their effectiveness in regression tasks \citep{garg2022what,vacareanu2024words}, where they have been found to approximate numerical mappings with surprisingly strong accuracy when prompted with in-context examples.

The intersection of LLMs and regression is particularly important for the longstanding task of text regression, where unstructured language must be mapped reliably to numeric outputs \citep{bitvai2015nonlinear}. Traditional regression models often struggle with applications where crucial information lies in unstructured text, such as product descriptions or financial reports \citep{zhang2023describe, gu2024predicting}, requiring rich text understanding. This is crucial in domains like product pricing where heterogeneous features across categories (e.g., screen technology for televisions versus mileage for cars) make traditional unified feature representations inadequate for capturing category-specific dynamics.

Existing work on regression with LLMs has explored three main approaches to address these challenges: fine-tuning LLMs for specific numeric prediction tasks \citep{jacobs2024regression}, using LLM embeddings as features for downstream regression models \citep{imperial2021bert,tang2024understanding}, and leveraging in-context learning for zero-shot or few-shot numeric estimation \citep{vacareanu2024words}. However, these approaches, with the exception of a few \cite{gruver2023large,qiu2024efllm}, focus on point estimates, overlooking a key limitation: the inability to quantify uncertainty. Many real-world applications, such as price prediction, demand forecasting, financial risk assessment, and healthcare operation inherently require probabilistic outputs rather than single-value predictions \citep{arora2023probabilistic,qiu2024efllm,gurlek2024boosted}.
Probabilistic modeling is essential in these applications to capture uncertainty and normal variation, mitigate risks, and improve decision-making \citep{gu2024predicting}. Current work has neither explored probabilistic regression using LLMs in-depth nor evaluated the trade-offs between different LLM regression approaches in a single study.

This paper presents the first study of probabilistic regression using LLMs to encode unstructured text inputs, and take a step towards a systematic investigation of LLM-based regression methods.
We center our study on price prediction, a task that requires both nuanced interpretation of free-form text inputs and accurate distribution estimation.
Understanding the complete price distribution is essential in financial contexts where accurately modeling tail behavior is critical for effective risk management.

In sum, this paper makes three key contributions. First, we propose a novel LLM-based quantile regression approach that produces full distributions with strong calibration while maintaining sharp prediction intervals, and improves point estimation accuracy compared to traditional approaches. Qualitative analysis shows that our model produces well-calibrated distributions that adapt to different price ranges and uncertainties across datasets, with tighter distributions for standardized products and appropriately wider distributions for items with more price variability. Second, we systematically compare different LLM architectures (decoder-only vs. encoder-only vs. traditional ML on text embeddings vs. in-context learning), multiple loss functions (squared error vs. pinball), and various data scales, while investigating training data contamination. We show that fine-tuning decoder models (e.g., Mistral-7B) outperforms other approaches. Our results confirm that model size, data scaling, and clean training sets all play critical roles in robust, generalizable LLM-based probabilistic regression. Third, we examine the application of LLMs in data cleaning for price estimation tasks. We demonstrate that LLM-guided label correction achieves comparable accuracy to human labeling without introducing systematic bias, and we make available three curated datasets (Amazon products, Craigslist used cars, and used boats) with standardized data splits.

\section{Related Work}
\label{sec:related-work}

\paragraph{Distributional, Quantile and Text Regression:}

Quantile regression, introduced by \citet{koenker1978regression}, extends beyond traditional pointwise prediction methods by characterizing the entire conditional distribution of the target variable through estimation of conditional quantiles at different probability levels \citep{kneib2023rage}. Unlike ordinary least squares regression which minimizes squared errors, quantile regression uses the \textit{pinball loss}, making it robust to outliers and capable of capturing heterogeneous effects across the distribution. This approach is especially valuable in healthcare operations, finance, and economics \citep{arora2023probabilistic,gurlek2024boosted,dichev2023estimating,gu2024predicting}, e.g. where modeling the full distribution helps capture both typical and extreme valuations.
Text Regression is a natural language processing task involving predicting continuous numerical values from unstructured text input with applications in domains like financial forecasting, election prediction, and box office revenue estimation \citep{bitvai2015nonlinear,dereli2019convolutional}. %

\paragraph{LLM-based Distribution Estimation:} Recent work has begun exploring LLMs' capabilities for distributional prediction tasks. \citet{gruver2023large} demonstrated probabilistic forecasting by mapping LLM token predictions to continuous distributions, while \citet{qiu2024efllm} developed a fine-tuned LLM that outputs discretized probability ranges for energy forecasting. However, these approaches focus primarily on structured numerical sequences rather than deriving distributions from unstructured text inputs, relying on either zero-shot prompting with specialized number formatting or domain-specific fine-tuning with predefined output ranges.
Our work addresses the broader challenge of predicting full probability distributions directly from unstructured text through quantile regression. 
We explore and compare several approaches for text-to-distribution models: (1) computing LLM embeddings then feeding them into a separate quantile prediction model, (2) extracting embeddings and approximating distributions using outcomes of ``neighboring'' embeddings in training data, and (3) our proposed approach of attaching multi-quantile heads directly to the LLM's last hidden layer and fine-tuning the entire architecture end-to-end with smoothed pinball loss. To our knowledge, this is the first paper to explore price prediction as a text-to-distribution NLP task rather than as a point value prediction task, with our approach allowing the LLM's representation layers to adapt in capturing distribution-relevant features.

\paragraph{Regression with LLMs Embeddings:}
A prevalent approach involves using pre-trained LLMs to generate text embeddings for downstream regression tasks \citep{imperial2021bert, gu2024predicting}. \citet{tang2024understanding} provide evidence that LLM embeddings maintain strong regression performance even as input dimensionality increases, where traditional feature engineering methods typically fail.

\paragraph{In-Context Learning for Regression:}
Recent research reveals the surprising capability of LLMs like GPT-4 and Claude to perform regression through in-context learning \citep{garg2022what, vacareanu2024words}. 
Their work shows that regression accuracy generally improves with the number of in-context examples provided. In the domain of real estate, \citet{chen2024predicting} confirm this behavior for price prediction tasks. \citet{lukasik2024regression} further advance this direction with Regression-Aware Inference with LLMs (RAIL), enhancing zero-shot numeric prediction through optimized decoding strategies. Their approach demonstrates that careful calibration of sampling parameters can significantly improve regression performance without requiring model fine-tuning.

\paragraph{Fine-Tuning LLMs for Regression:}
Recent studies demonstrate the effectiveness of fine-tuning LLMs for regression tasks across diverse domains. \citet{jacobs2024regression} fine-tune a LLaMA-based model, achieving performance comparable to specialized domain models in chemical property prediction. \citet{zhang2023describe} show that fine-tuned BERT-based models can effectively predict house prices from unstructured property descriptions, outperforming baselines that rely solely on structured features. \citet{song2024omnipred} present a framework that converts various input formats into text and fine-tunes LLMs as universal end-to-end regressors, demonstrating strong cross-domain performance.

While these studies demonstrate LLMs' capabilities for regression, two gaps exist in the literature.
Most critically, existing work largely overlooks probabilistic regression techniques that could leverage LLMs' rich understanding of textual data.
Additionally, prior work lacks unified comparisons between fine-tuning, embedding-based methods, and in-context learning under similar conditions.
We primarily address the probabilistic gap by introducing a novel method for quantile regression that enable uncertainty-aware predictions from language models, while also providing a comparative analysis of different LLM-based regression approaches.

\section{Price Estimation Datasets}
\label{sec:data}
We experiment upon three price prediction datasets from different domains publicly available for research use: Amazon Products~\cite{ni2019justifying}, Craigslist Used Cars listings,\footnote{\tiny \url{https://kaggle.com/datasets/austinreese/craigslist-carstrucks-data}} and European Boat Sales.\footnote{\tiny \url{https://www.kaggle.com/datasets/karthikbhandary2/boat-sales}} Examples are provided in Table \ref{tab:data_format_examples}.
Initial manual inspection revealed numerous instances with erroneous prices, i.e., unreasonably %
high or low relative to the items being sold (examples are provided in \Cref{tab:bad_prices} of the Appendix). To address this, we employed the Claude-3.5-Sonnet LLM \cite{anthropic2024claude} in a zero-shot manner to identify and remove rows with incorrect prices from all dataset splits. Details of this process and our human evaluation ($>$94\% agreement between human and LLM judgments) verifying that it neither removed difficult instances nor introduced bias are provided in Figure \ref{fig:cleaningprompt} and Appendix \ref{subsec:cleanup-eval}. 
\Cref{tab:dataset_stats} presents the dataset distribution and number of samples removed %
for each dataset.%
\begin{table}[ht]
\centering
\small
\setlength{\tabcolsep}{3pt}
\begin{tabular}{lrrrr}
\hline
Dataset & Train & Val & Test & Removed \\
\hline
Amazon Products & 500K & 15K & 15K & ~100K \\
Craigslist Used Cars & 350K & 19K & 19K & ~10K \\
Boats & 8K & 450 & 450 & ~1K \\
\hline
\end{tabular}
\caption{Dataset Size Distribution}
\label{tab:dataset_stats}
\end{table}

\section{LLM-Based Quantile Regression}
\label{sec:methods}

\subsection{Problem Statement}
\begin{figure}[t]
    \centering
    \includegraphics[width=\columnwidth]{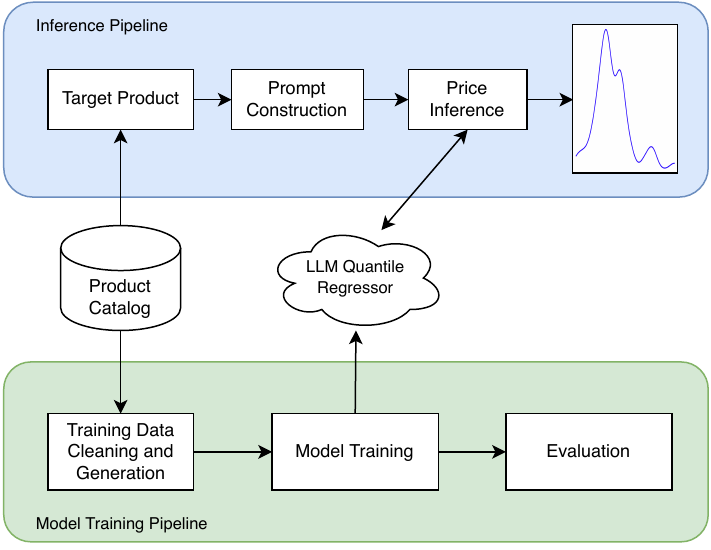}
    \caption{End-to-end training and inference pipelines of our proposed framework.} %
    \label{fig:pipeline}
\end{figure}

Given a random variable $X$ with realizations $X=\vx\in \mathcal{X}$ representing unstructured textual input (e.g., a product title or description) and other structured attributes, our goal is to predict the conditional distribution $F_{Y|X}(\cdot|\vx)$, where $y \in \reals$ is a numeric outcome, such as the price of a product.
More formally, we aim to learn a function $f(\cdot;\Theta): \mathcal{X} \to \mathcal{F}$ that maps inputs to conditional distributions, where $\mathcal{F}$ is the space of cumulative distribution functions on $\reals$. Here, $\Theta$ is a multi-dimensional parameter, such as the weights of an LLM. We represent these distributions through a vector of conditional quantiles $\vquant_{\vtau}(\vx)=(\hquant_{\tau_1}(\vx),\ldots,\hquant_{\tau_K}(\vx))$ for $K$, pre-specified quantiles $\vtau=(\tau_1, ..., \tau_K) \in (0,1)^K$. The optimal parameters $\Theta^*$, are learned by minimizing:
\[
\Theta^* = \arg\min_\Theta \E_{(\vx,y)\sim\mathcal{D}}[\loss(f(\vx;\Theta), y)]
\]
where $\mathcal{D}$ is the underlying data distribution and $\mathcal{L}$ is a proper loss function for probabilistic forecasts. Figure \ref{fig:pipeline} shows the end-to-end training and inference pipelines of our proposed approach for distributional price prediction using LLM-based quantile regression.

\subsection{Quantile Regression Head \& Pinball Loss}\label{subsec:quantile-head}

We propose adding a quantile regression head to both decoder-only and encoder-only LM architectures.
Let \(f(\cdot;\Theta)\) be the decoder-only LLM parameterized by \(\Theta\). It takes a tokenized version of the input text ($\vx$) with sequence length $T$, denoted by \(\tx = (x_{1}, \ldots, x_{T})\), and produces hidden states \(H \in \mathbb{R}^{T \times D}\) where $D$ is hidden state dimension. We then extract the final hidden state \(\vh_{T} = H[T,:]\) as a summary of the sequence.
For encoder models, $\vh_T$ is the [CLS] token representation.
We then replace the language modeling head with a quantile regression head \(g(\cdot;\phi)\), to predict \(K\) quantiles \(\hat\vquant = (\hquant_{\tau_1},\ldots,\hquant_{\tau_K})\) given \(\mathbf{h}_{T}\), with additional refinements in  \cref{sec:quantile_monotonicity}. 
Therefore, our model structure is:
\begin{align}
\tx&=\text{tokenized }\vx\,,\\
H &= f(\tx; \Theta) \in \mathbb{R}^{T \times D}\,, \\
\vh_T &= H[T, :]\,, \\
\hat{\vquant} &= g\!\bigl(\vh_T; \phi\bigr) \in \mathbb{R}^{K}\,.\label{eq:output-of-llm}
\end{align}

The pinball loss, also known as the quantile loss, enables asymmetric penalization of over- and under-predictions. For predicted quantiles $\hat{\vquant}$ obtained in \eqref{eq:output-of-llm}, given quantile levels $\vtau$, and ground truth $y$, we implement the pinball loss as
$
\mathcal{L}_{\vtau}(\hat{\vquant},y) = (1/K)\sum_{k=1}^K \loss_{\tau_k}(\hquant_{\tau_k},y)
$, where,
\begin{equation}
\loss_\tau(\hquant_\tau,y) = \tau(y - \hquant_\tau) + \relu(\hquant_\tau - y)\,.
\end{equation}
For improved optimization stability, we employ a smoothed variant:
\begin{equation*}
\loss_\tau^\alpha(\hquant_\tau,y) = \tau(y - \hquant_\tau) + \alpha \cdot \softplus_\alpha(\hquant_\tau-y),
\end{equation*}
where $\softplus_\alpha(x) = \alpha\log(1 + e^{x/\alpha})$ provides a differentiable approximation to $\relu$ as $\alpha \to 0^+$.

\section{Experimental Setup}

\paragraph{Quantile Levels and Point Prediction:} 
For all models that predict distributions, we take $K=200$ and divide the interval $(0,1)$ into $K$ equal-length sub-intervals to obtain $\vtau$. We discuss the impact of varying $K$ and the smoothing parameter $\alpha$ in Appendix~\ref{app:metrics_and_definitions}. 
We use models that produce a distribution both for generating probabilistic outputs and for point predictions. In the latter case, we take the predicted quantile at $\tau=0.5$ as the point estimate. Additionally, as baselines, we include models trained solely with traditional squared error loss, using their direct predictions for comparison.

\noindent \textbf{Baselines Using LLM Embeddings: }
Text features (title, description, attributes) are concatenated with appropriate field markers and converted to embeddings using the Qwen2-7B-instruct embedding model \citep{chu2024qwen2}. All baseline models using these embeddings are denoted by the ``Qwen-7B-Emb'' prefix. These embeddings serve as input features for five models: Ridge Regression and XGBoost for point estimation, Quantile Regression (with two hidden layers) for distribution prediction, trained on log-transformed target,\footnote{In all three data sets since the target was price, we used its log-transformed prices to handle the wide range of values in our datasets during training.} and two nearest neighbor-based distribution prediction approaches. The first nearest neighbor model (kNN) predicts distributions by using the empirical distribution of target values from selected neighbors in the training set, while the second variant employs a radius-based selection criterion (rkNN) with a minimum neighbor requirement. The rationale for rkNN is to ensure more accurate empirical distribution among the neighbors, compared to kNN. All hyperparameters are selected using 5-fold cross-validation.

\noindent \textbf{Fine-tuned LMs with Quantile Head: }We fine-tune Mistral-7B \cite{mistral2023mistral7b}, Phi-3B \cite{abdin2024phi3}, Qwen-500M \cite{bai2023qwen} and XLM-RoBERTa \cite{conneau2020unsupervised} models with the quantile regression head described in Section \ref{sec:methods}. %

\noindent \textbf{In-context Learning: }We evaluate two state-of-the-art LLMs, Claude-3.5-Sonnet and Nova Pro \cite{anthropic2024claude,aws2024nova}, both zero-shot and few-shot. For few-shot learning, we implement three example selection strategies: (i) random sampling; (ii) category-based stratified sampling and (iii) similar item sampling based on cosine similarity of Qwen2-7B embeddings. The latter two 
leverage domain similarity for potentially better price estimation (prompts in Figure~\ref{fig:fewshotprompt}). %

\paragraph{Evaluation Metrics:}
We use two sets of metrics. The first set evaluates point price estimates and includes: (i) \textbf{Mean Absolute Percentage Error (MAPE)}, (ii) \textbf{Weighted Absolute Percentage Error (WAPE)}, with weight = 1 and (iii) \textbf{Mean Percentage Error (MPE)}. 

The second set of metrics measures the distributional quality of the predicted quantiles %
$\hat{\vquant}_{\vtau}(\vx_i)=(\hquant_{\tau_1}(\vx_i) \le \dots \le \hquant_{\tau_K}(\vx_i))$ for each input $\vx_i$.
These metrics include:

\textbf{(i) Calibration Error (CE)}
measures how well predicted quantiles match their theoretical coverage.
$\mathrm{CE}
=
(1/K)\sum_{k=1}^{K}
|\widehat{\mathrm{coverage}}(\tau_k) - \tau_k|
$, 
where $\widehat{\mathrm{coverage}}(\tau_k)$ is the empirical fraction of true values in the test set, below the $\tau_k$ quantile.

\textbf{(ii) Continuous Ranked Probability Skill Score (CRPSS)}
measures the integrated squared difference between predicted and true cumulative distribution functions.

\textbf{(iii) Relative Confidence Interval Width (RCIW)}
measures the average width or tightness of predicted intervals relative to the true value.

For each metric, we report 95\% confidence intervals with bootstrap resampling (1000 iterations):
$\text{CI}_{95\%}(M) = [\hat{M}_{(0.025)}, \hat{M}_{(0.975)}]$
where $\hat{M}_{(q)}$ denotes the $q$-th quantile of the bootstrap distribution of metric $M$. 
Further details about all metrics and the training process are presented in Appendix \ref{app:metrics_and_definitions}.

\section{Results}
\label{sec:results}

\begin{table*}[!th]
\centering
\small
\setlength{\tabcolsep}{4pt}
\resizebox{\textwidth}{!}{%
\begin{tabular}{llrrrrrrrr}
\toprule
\multirow{2}{*}{\textbf{Dataset}} & \multirow{2}{*}{\textbf{Model}} & \multicolumn{2}{c}{MAPE (\%) \textcolor{red}{$\downarrow$}} & \multicolumn{2}{c}{MPE (\%) \textcolor{red}{$\downarrow$}} & \multicolumn{2}{c}{WAPE (\%) \textcolor{red}{$\downarrow$}} \\
\cmidrule(lr){3-4} \cmidrule(lr){5-6} \cmidrule(lr){7-8}
& & Value & 95\% CI & Value & 95\% CI & Value & 95\% CI \\
\midrule
\multirow{7}{5em}{Amazon Products} 
& Mistral-7B-Point & 20.81 & [20.13, 21.22] & -3.40 & [\phantom{0}-3.53, \phantom{0}-3.30] & 22.81 & [22.45, 24.67]  \\
& Mistral-7B-Quantile & \textbf{16.86} & [16.15, 17.71] & \textbf{-0.88} & [\phantom{0}-1.23, \phantom{0}-0.55] & \textbf{18.32} & [17.83, 18.83] \\
& XLM-R Base-Quantile & 41.99 & [40.34, 43.86] & -21.73 & [-23.72, -20.00] & 40.27 & [39.09, 41.45] \\
& XLM-R Large-Quantile & 36.52 & [34.64, 38.49] & -15.18 & [-17.22, -13.22] & 37.51 & [36.20, 38.74] \\
& Qwen-500M-Quantile & 39.19& [38.01, 40.36] & -6.33 & [\phantom{0}-7.73, \phantom{0}-5.07]& 43.15 & [42.07, 44.10]\\
& Phi-3B-Quantile & 34.17 & [33.14, 35.27] & -5.41 & [\phantom{0}-6.65, \phantom{0}-4.17] & 38.17 & [37.11, 39.29] \\
& Qwen-7B-Emb+Ridge & 58.97 & [57.78, 60.26] & 30.36 & [\phantom{-}32.02, -29.02] & 52.72 & [51.93, 53.41] \\
& Qwen-7B-Emb+XGBoost & 63.16 & [62.22, 64.30] & -32.57 & [-33.98, -31.50] & 58.01 & [57.27, 58.88] \\
& Qwen-7B-Emb+Quantile & 77.97 & [76.89, 79.10] & -27.44 & [-28.99, -25.88] & 76.3 & [75.69, 76.99] \\
& Qwen-7B-Emb+kNN-Quantile & 46.86 & [45.88, 47.83] & -10.53 & [-11.68, \phantom{0}-9.27] & 54.05 & [52.95, 55.04] \\
& Qwen-7B-Emb+RkNN-Quantile & 42.68 & [41.66, 43.90] & -10.33 & [-11.65, \phantom{0}-9.06] &48.03 &[46.88, 49.21]  \\ 
& \shortstack[l]{Claude-3.5-Sonnet \\ (512 category-based shots)} & 38.50 & [36.70, 39.10]  & 14.32   & [14.29, 14.41]  & 41.40  & [40.20, 42.16]  \\ %
& \shortstack[l]{Nova-Pro \\ (512 category-based shots)} & 43.77 & [40.78, 45.01]  &  19.12  & [18.79, 19.81]  & 48.13  & [46.21, 49.33]  \\ 
\midrule
\multirow{7}{*}{Used Cars} 
& Mistral-7B-Point & 9.76 & [\phantom{0}9.25, 10.67] & -5.40 & [\phantom{0}-5.89, \phantom{0}-4.01] & 12.79 & [12.65, 13.32] \\
& Mistral-7B-Quantile & \textbf{6.30} & [\phantom{0}6.06, \phantom{0}6.95] & \textbf{0.19} & [\phantom{-0}0.05, \phantom{-0}0.31] & \textbf{5.40} & [\phantom{0}5.29, \phantom{0}5.51] \\
& XLM-R Base-Quantile & 11.45 & [10.68, 12.44] & -5.71 & [\phantom{0}-6.62, \phantom{0}-4.87] & 8.89 & [\phantom{0}8.62, \phantom{0}9.23] \\
& XLM-R Large-Quantile & 12.84 & [12.41, 13.37] & -9.70 & [-10.23, \phantom{0}-9.24] & 10.46 & [10.22, 10.74] \\
& Qwen-500M-Quantile & 23.49 & [20.61, 26.71] & -4.56 & [\phantom{0}-8.03, \phantom{0}-1.48] & 15.93 & [15.48, 16.40]\\
& Phi-3B-Quantile & 52.79 & [51.83, 53.89] & 47.09 & [\phantom{-}45.90, \phantom{-}48.14] & 74.91 & [74.50, 75.32] \\
& Qwen-7B-Emb+Ridge & 40.46 & [37.95, 43.42] & -18.04 & [-21.12, -15.44] & 23.04 & [22.49, 23.41] \\
& Qwen-7B-Emb+XGBoost & 39.70 & [38.10, 41.75] & -16.09 & [\phantom{-}17.96, -14.41] & 26.13 & [25.80, 26.60] \\
& Qwen-7B-Emb+Quantile & 235.92 & [221.57, 249.55] & -192.67 & [-206.41, -178.24] & 58.41 & [57.85, 58.85] \\
& Qwen-7B-Emb+kNN-Quantile & 79.72 & [73.18, 86.87] & -59.39 & [-66.09, -52.59] & 26.81 & [26.33, 27.29] \\
& Qwen-7B-Emb+RkNN-Quantile &58.18  &[52.46, 64.02]  &-40.92  & [-46.86, -35.39] &21.37 &[20.93, 21.84]  \\
& \shortstack[l]{Claude-3.5-Sonnet \\ (2048 random shots)} & 275.00 & [269.12, 280.09] & 189.19 & [\phantom{-}175.21, \phantom{-}195.62] & 53.34 & [50.78, 56.09]  \\ 
& \shortstack[l]{Nova-Pro \\ (1024 random shots)} & 219.67 & [167.42, 231.91] & 173.07 & [\phantom{-}156.12, \phantom{-}189.07] & 46.44 & [42.13, 48.71]  \\ %
\midrule
\multirow{7}{*}{Boats} 
& Mistral-7B-Point & 24.01 & [23.82, 24.29] & 4.10 & [2.30, \phantom{00}7.45] & 25.82 & [24.20, 27.39] &   \\
& Mistral-7B-Quantile & \textbf{21.20} & [20.50, 23.39] & 2.19 & [1.59, \phantom{00}6.65] & \textbf{23.96} & [20.68, 27.69] \\
& XLM-R Base-Quantile & 22.17 & [20.26, 24.47] & \textbf{0.58} & [-2.69, \phantom{00}3.52] & 23.59 & [20.12, 26.78] \\
& XLM-R Large-Quantile & 22.67 & [20.85, 24.55] & -4.51 & [-7.43, \phantom{--}-1.78] & 31.05 & [24.31, 37.99] \\
& Qwen-500M-Quantile & 62.27 & [56.49, 69.12] & 16.98 & [8.6, \phantom{0}24.73] & 77.23 & [73.20, 80.75] \\
& Phi-3B-Quantile & 73.83 & [71.45, 76.02] & 72.89 & [\phantom{-}70.32, \phantom{-}75.31] & 93.64 & [92.41, 94.64] \\
& Qwen-7B-Emb+Ridge & 30.77 & [28.06, 33.89] & -7.52 & [-11.85, \phantom{0}-3.81] & 28.77 & [24.57, 33.38] \\
& Qwen-7B-Emb+XGBoost & 44.56 & [40.12, 49.19] & -12.84 & [-17.86, \phantom{0}-6.93] & 42.35 & [38.02, 46.85] \\
& Qwen-7B-Emb+Quantile & 131.03 & [110.78, 158.77] & -67.61 & [-99.21, -44.88] & 82.21 & [78.75, 84.98] \\
& Qwen-7B-Emb+kNN-Quantile & 77.68 & [67.39, 88.29] & -32.36 & [-43.49, -20.56] & 63.39 & [58.06, 67.78] \\
& Qwen-7B-Emb+RkNN-Quantile &70.96  & [61.86, 80.72] & -28.67 & [-39.61, -18.10] &56.80 &[51.81, 61.44]  \\
& \shortstack[l]{Claude-3.5-Sonnet \\ (2048 random shots)} & 30.00 & [28.97, 31.28] & 17.32  & [15.16, 19.23] & 29.36 & [26.16, 30.09]  \\ %
& \shortstack[l]{Nova-Pro \\ (2048 random shots)} & 61.01 & [55.54, 64.76]  &  23.22  & [21.16, 25.91]  & 48.79  & [45.03, 50.71]  \\ 
\bottomrule
\end{tabular}
}
\begin{tablenotes}
\small
\item \textbf{Bold} values indicate best performance for each metric and dataset. The \textcolor{red}{$\downarrow$} indicates that lower metric values are better.
\end{tablenotes}
\caption{Model point-estimate performance comparison, using median as the point estimate for quantile regression models. For the few-shot Claude-3.5 and Nova-Pro LLMs, we only show the optimal few shot example selection strategy and the corresponding number of shots that gave the best results.}%
\label{tab:model_comparison}
\end{table*}

\begin{table*}[!bht]
\centering
\small
\setlength{\tabcolsep}{4pt}
\begin{tabular}{llrrrrrrrr}
\toprule
\multirow{2}{*}{\textbf{Dataset}} & \multirow{2}{*}{\textbf{Model}} & \multicolumn{2}{c}{CE \textcolor{red}{$\downarrow$}} & \multicolumn{2}{c}{CRPSS \textcolor{blue}{$\uparrow$}} & \multicolumn{2}{c}{RCIW@95\%CI \textcolor{red}{$\downarrow$}} \\
\cmidrule(lr){3-4} \cmidrule(lr){5-6} \cmidrule(lr){7-8}
& & Value & 95\% CI & Value & 95\% CI & Value & 95\% CI \\
\midrule
\multirow{7}{*}{Amazon Products} 
& Mistral-7B-Quantile & 0.042  & [0.039, 0.044] & \textbf{0.75}  & [0.74, 0.76] & \textbf{0.92}  & [0.92, 0.93] \\ 
& XLM-R Base & 0.060  & [0.057, 0.064] & 0.49  & [0.48, 0.51] & 2.03  & [1.99, 2.09] \\ 
& XLM-R Large & 0.040  & [0.037, 0.043] & 0.53  & [0.51, 0.55] & 1.52  & [1.48, 1.57] \\ 
& Qwen-500M-Quantile & 0.055  & [0.051, 0.059] & 0.47  & [0.46, 0.49] & 2.89  & [2.83, 3.00] \\ 
& Phi-3B-Quantile & 0.041  & [0.036, 0.046] & 0.53  & [0.52, 0.54] & 2.33  & [2.27, 2.38] \\ 
& Qwen-7B-Emb+Quantile & 0.045  & [0.042, 0.048] & 0.03  & [0.01, 0.04] & 16.76  & [16.59, 16.92] \\ 
& Qwen-7B-Emb+kNN-Quantile & \textbf{0.01}  & [0.006, 0.013] & 0.34  & [0.31, 0.37] & 6.14  & [6.05, 6.26] \\ 
& Qwen-7B-Emb+RkNN-Quantile & \textbf{0.01}  & [0.007, 0.012]  & 0.42  & [0.39, 0.44] & 6.12  & [6.00, 6.29] \\
\midrule
\multirow{7}{*}{Used Cars} 
& Mistral-7B-Quantile & 0.054  & [0.051, 0.055] & \textbf{0.92}  & [0.91, 0.92] & \textbf{0.20}  & [0.20, 0.21] \\ 
& XLM-R Base & 0.157  & [0.155, 0.159] & 0.80  & [0.79, 0.81] & 1.02  & [1.01, 1.03] \\ 
& XLM-R Large & 0.185  & [0.183, 0.187] & 0.80  & [0.79, 0.81] & 1.01  & [1.00, 1.01] \\ 
& Qwen-500M-Quantile & 0.160  & [0.158, 0.162] & 0.66  & [0.65, 0.66] & 4.70  & [3.76, 5.50] \\ 
& Phi-3B-Quantile & 0.395  & [0.393, 0.397] & 0.04  & [0.03, 0.05] & 0.99  & [0.96, 1.04] \\ 
& Qwen-7B-Emb+Quantile & \textbf{0.020}  & [0.018, 0.022] & 0.01  & [0.01, 0.02] & 13.41  & [12.84, 14.45] \\ 
& Qwen-7B-Emb+kNN-Quantile & 0.024  & [0.020, 0.028] & 0.53  & [0.52, 0.53] & 3.90  & [3.75, 4.09] \\ 
& Qwen-7B-Emb+RkNN-Quantile & 0.022  & [0.019, 0.026] & 0.62  & [0.63, 0.64] & 2.64  & [2.54, 2.80] \\
\midrule
\multirow{7}{*}{Boats} 
& Mistral-7B-Quantile & 0.076  & [0.070, 0.084] & \textbf{0.73}  & [0.67, 0.77] & 1.28  & [1.23, 1.35] \\ 
& XLM-R Base & 0.047  & [0.028, 0.066] & 0.73  & [0.70, 0.77] & 1.54  & [1.50, 1.59] \\ 
& XLM-R Large & 0.042  & [0.030, 0.051] & 0.59  & [0.45, 0.69] & 1.68  & [1.65, 1.72] \\ 
& Qwen-500M-Quantile & 0.257  & [0.237, 0.275] & 0.18  & [0.03, 0.44] & 1.24  & [1.17, 1.34] \\ 
& Phi-3B-Quantile & 0.453  & [0.445, 0.461] & 0.21  & [0.13, 0.35] & \textbf{0.68}  & [0.62, 0.74] \\ & Qwen-7B-Emb+Quantile & 0.034  & [0.014, 0.056] & 0.20  & [0.09, 0.23] & 33.37  & [30.03, 36.55] \\ 
& Qwen-7B-Emb+kNN-Q & \textbf{0.021}  & [0.011, 0.036] & 0.28  & [0.19, 0.38] & 11.33  & [10.41, 12.25] \\ 
& Qwen-7B-Emb+RkNN-Q & 0.025  & [0.013, 0.042] & 0.39  & [0.30, 0.46] & 7.87  & [7.03, 8.61] \\
\bottomrule
\end{tabular}
\begin{tablenotes}
\small
\item \textbf{Bold} values indicate best performance. \textcolor{red}{$\downarrow$} indicates that lower metric values are better, and \textcolor{blue}{$\uparrow$} indicates that higher are better.
\end{tablenotes}
\caption{Model distribution prediction performance comparison across models and datasets. CE measures how well predicted quantiles match their theoretical coverage, CRPSS evaluates the probabilistic prediction quality relative to a reference, and RCIW measures the sharpness of the distribution prediction intervals.}
\label{tab:results_distribution}
\end{table*}

\subsection{Point Regression Results}
\Cref{tab:model_comparison} lists our main point regression results, comparing all models across three datasets.

\paragraph{Fine-tuned LLMs Outperform Traditional Models.}
The fine-tuned Mistral-7B-Quantile model notably %
outperforms other approaches across all datasets. For the Amazon Products dataset, Mistral-7B achieves a MAPE of 16.86\%, substantially lower than the best traditional baselines (Qwen7B-Emb+RkNN-Q) at 42.68\%. This pattern is particularly pronounced in the Used Cars dataset, where Mistral-7B's MAPE of 6.3\% represents an order of magnitude improvement over traditional approaches, which show MAPEs up to 235\%. This highlights the importance of rich text understanding for price regression.
The MPE results indicate that traditional approaches tend to systematically underestimate prices, with negative biases ranging from -24\% to -135\% across datasets. In contrast, Mistral-7B shows minimal systematic bias, with MPE values close to zero: -0.88\% for Amazon Products and 0.185\% for Used Cars.

\paragraph{Better Estimates via Quantile Regression.}
Comparing the best model (Mistral-7B-Quantile) to a version with a point regression head (Mistral-7B-Point) shows a substantial improvement in all metrics, confirming that the median of the quantile regression distribution is a better estimate than pointwise regression.

\paragraph{Decoder vs. Encoder Models.} Comparing encoder-only (XLM-RoBERTa) and decoder-only (Mistral-7B, Phi-3B, Qwen-500M) architectures provides nuanced insights. While the best-performing model overall is the larger decoder-only Mistral-7B, achieving notably lower MAPE across all datasets (16.9\% for Amazon Products, 6.3\% for Used Cars, and 21.2\% for Boats), this advantage is not uniformly supported by size-matched comparisons. Specifically, the encoder-based XLM-R-Large consistently outperforms the similarly sized decoder-only Qwen-500M across all datasets and even surpasses the larger Phi-3B on two datasets (Used Cars and Boats). Additionally, the smaller encoder-based XLM-R-Base remains competitive with Qwen-500M. The narrowing performance gap between Mistral-7B and XLM-R-Large on Boats data set (21.2\% vs. 22.7\%) suggests that model performance is influenced by a combination of factors beyond architecture alone, such as model size, data, and training methodology. Given that our primary objective is identifying the strongest overall text-to-distribution forecasting model, we defer a more thorough investigation that isolates and quantifies the precise architectural effects on point prediction accuracy to future research.

\begin{figure}[t]
    \centering
    \includegraphics[width=\columnwidth]{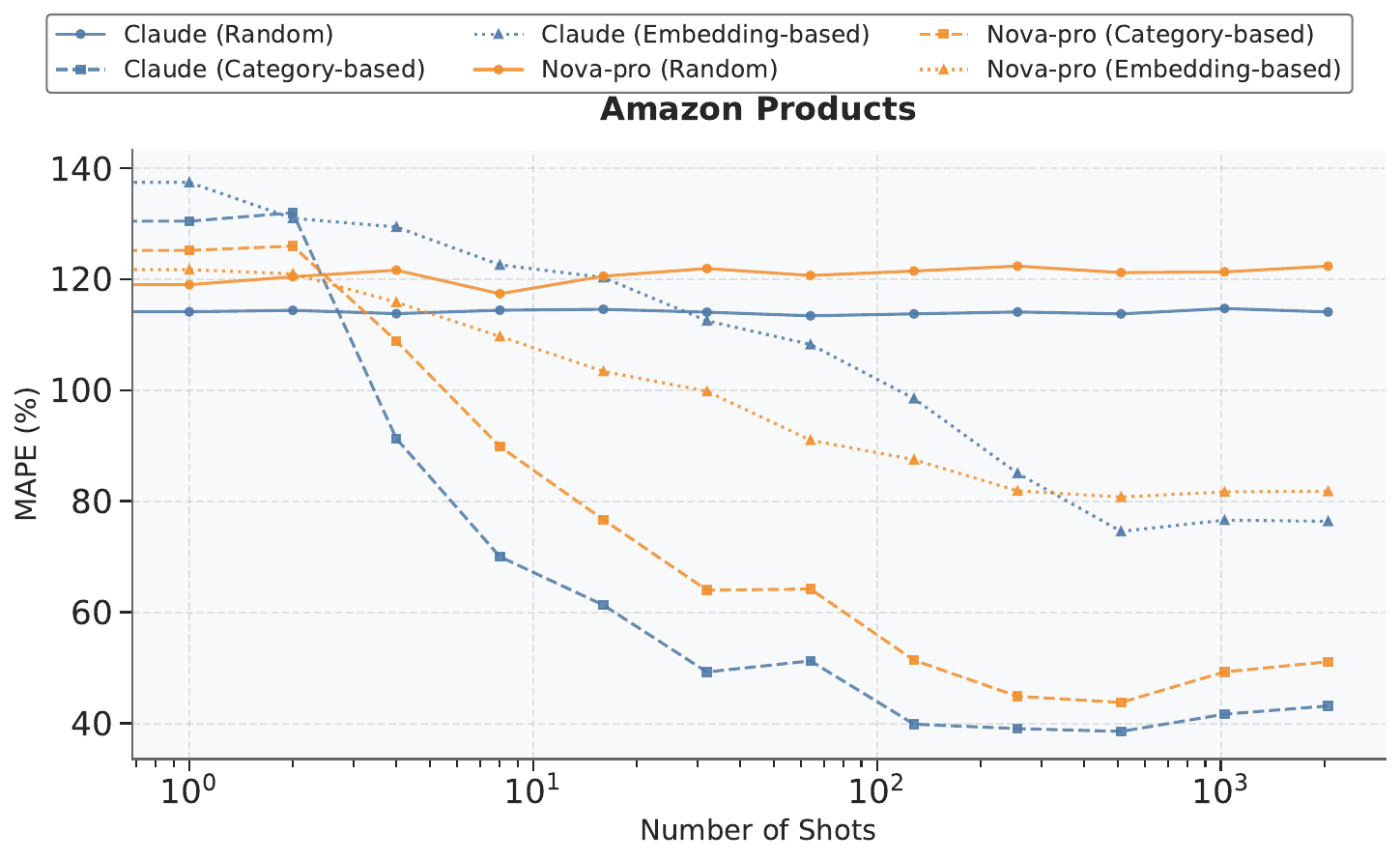}
    \caption{Few-shot learning performance of Claude-3.5-Sonnet and Nova-Pro LLMs on Amazon Products data.} %
    \label{fig:few_shot_scaling}
\end{figure}

\paragraph{Few-shot learning Underperforms.}
\Cref{fig:few_shot_scaling} shows the results of few-shot learning approaches across our three datasets. %
Even with the best-performing category-based sampling strategy and optimal shot count, both Claude and Nova-pro lag behind fine-tuned Mistral-7B by more than 15\% for Amazon Products and Used Boats, and by more than 200\% for Used Cars. %
Our experiments also reveal that increasing the number of examples beyond a certain point %
starts degrading model performance. 
This substantial performance gap indicates that for precise price prediction, fine-tuning yields considerably better results than carefully crafted few-shot approaches. It would appear that the complex relationships between rich textual data and prices require more thorough model adaptation than what can be achieved through in-context learning.

\subsection{Distributional Regression Results}

\begin{figure*}[ht]
    \centering
    \includegraphics[width=.32\textwidth]{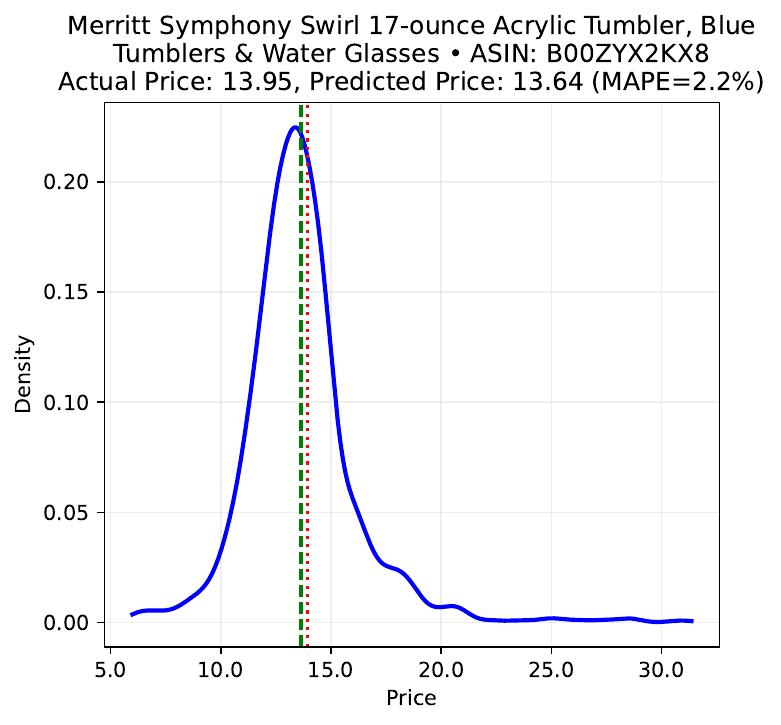}
    \includegraphics[width=.32\textwidth]{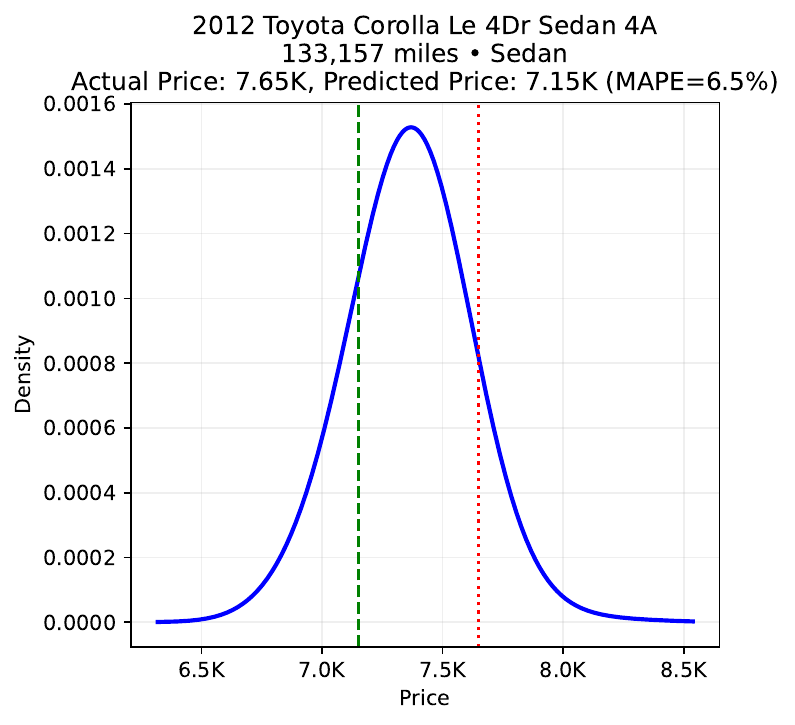}
    \includegraphics[width=.32\textwidth]{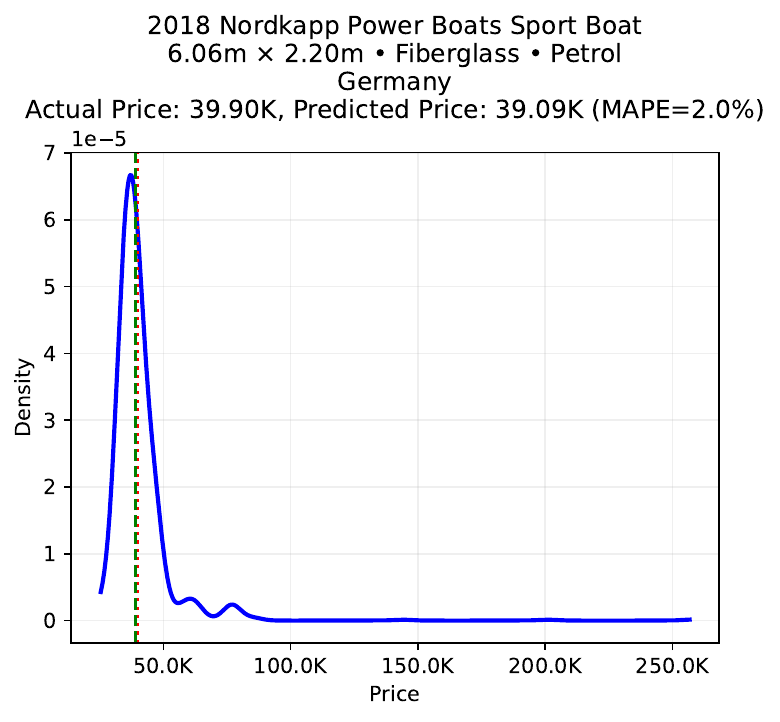}    
    \includegraphics[width=.32\textwidth]{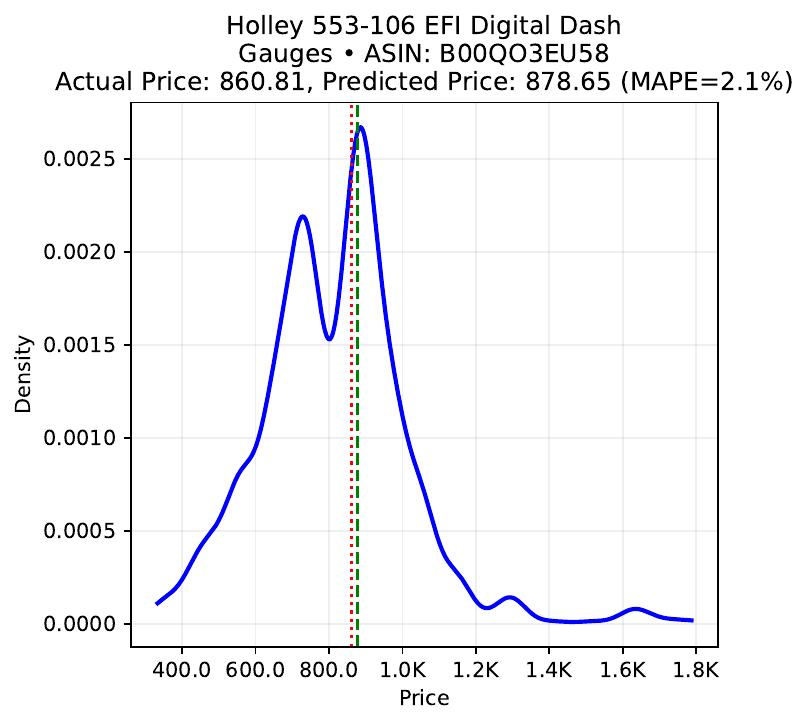}
    \includegraphics[width=.32\textwidth]{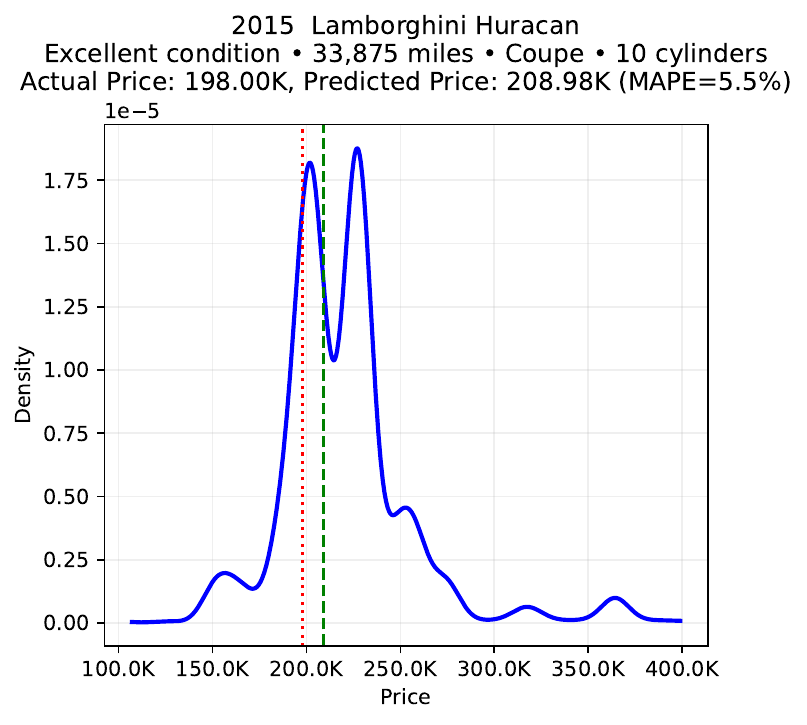}
    \includegraphics[width=.32\textwidth]{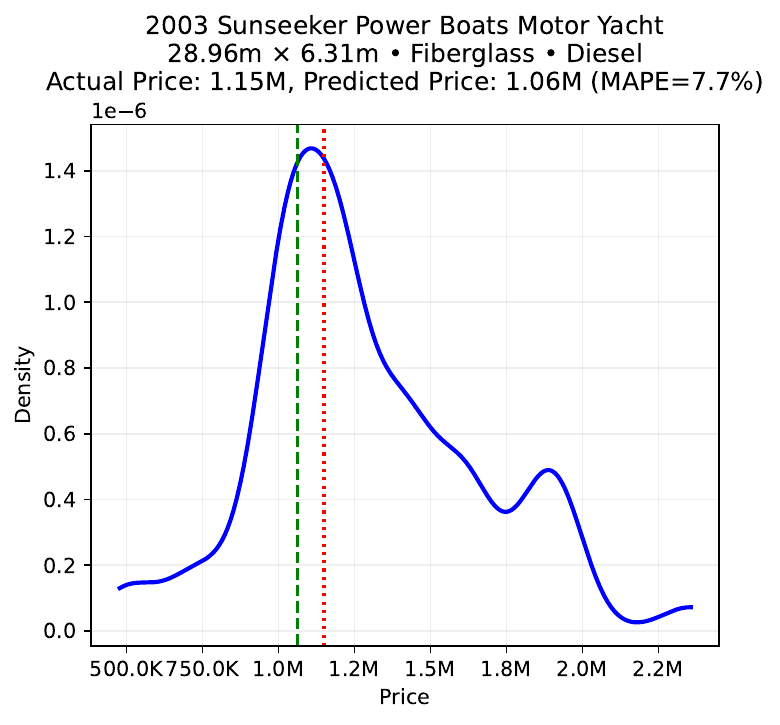}
    
    \caption{Probability density distribution of the prices predicted by the Mistral-7B-Quantile model across different datasets (\textcolor{blue}{blue} curve). Each x-axis has a different scale. The {\color{red}red} dotted line represents the ground truth price while the \textcolor{mydarkgreen}{green} dashed line is the predicted median price. As demonstrated, the model captures different distribution shapes including unimodal (top row),  bimodal (bottom row), and right-skewed (right) distributions.}
    \label{fig:pred_dist_plots}
\end{figure*}

\Cref{tab:results_distribution} lists our main distributional regression results, comparing various decoder and encoder models, and embedding baselines.

\paragraph{Larger Fine-tuned LMs achieve the best results.}
As earlier, the fine-tuned Mistral-7B-Quantile model is a consistently strong performer across all distributional metrics and datasets. It achieves the best CRPSS scores ranging between 0.73-0.92 on all three datasets, indicating the high quality of the model's probabilistic predictions relative to the references. 
Mistral-7B also achieves the best RCIW score on Amazon Products and Used Cars data, and a competitive score on the Boats data, indicating sharp distributions and generally precise prediction confidence intervals. Overall, Mistral-7B is much more consistent and maintains a better balance across different metrics compared to the smaller models like Phi, Qwen-500M or XLM-RoBERTa. 
Both Mistral-7B and Qwen-7B embedding based variants show very low CE scores on all datasets, indicating that the predicted probabilities match their theoretical coverage very well.

\paragraph{LLMs produce Better Calibrated Distributions.}

Mistral-7B-Quantile model demonstrates strong calibration (low CE between 0.04-0.07) while maintaining better confidence intervals, suggesting that fine-tuned LLMs are inherently better at producing well-calibrated probability distributions. Qwen-7B embedding variants also achieve very low calibration errors (CE of 0.01 on Amazon Products and 0.02 on Used Cars), significantly outperforming smaller models like XLM-RoBERTa which has a CE of 0.04-0.06 on Amazon Products and 0.157-0.185 on Used Cars. The RCIW patterns also reveal interesting trade-offs. While embedding-based Qwen-7B variants achieve excellent calibration, they produce much wider confidence intervals with RCIW between 6.12-16.76 on Amazon Products. However, the Mistral-7B model fine-tuned for quantile regression can achieve both sharp predictions and good calibration, evidenced by its optimal RCIW scores while also maintaining high CRPSS. 

\paragraph{Larger Data Leads to Better Distributions.}

Most models achieve overall better and more consistent distributional metric scores on the larger Amazon Products and Used Cars datasets compared to the much smaller Boats data. Mistral-7B-Quantile achieves tighter confidence intervals on Used Cars with an RCIW of 0.2 compared to its RCIW of 1.28 on the Boats data.
Wider confidence intervals and more variable model performance on the Boats dataset highlight the detrimental impact of smaller sample sizes on probabilistic predictions.

\paragraph{Qualitative Analysis of Distributions.}

We show in Figure~\ref{fig:pred_dist_plots} the predicted probability distribution function of the prices by our fine-tuned Mistral-7B-Quantile model, smoothed using a Gaussian kernel. We show examples from all three datasets having different MAPE values (additional examples are provided in Section~\ref{sec:addnl_dist_results} of the appendix). 
In both the Amazon Products examples, the predicted median and actual ground truth prices are very close to each other, with the largest mode of the distribution centered around the ground truth price. 
We see wider distributions spanning larger price ranges for the higher priced Used Cars datasets, with the most distribution width and price uncertainty in the Boats datasets, possibly due to a greater price variability in this domain or a smaller training dataset.

\begin{figure}[ht]
    \centering    \includegraphics[width=\columnwidth]{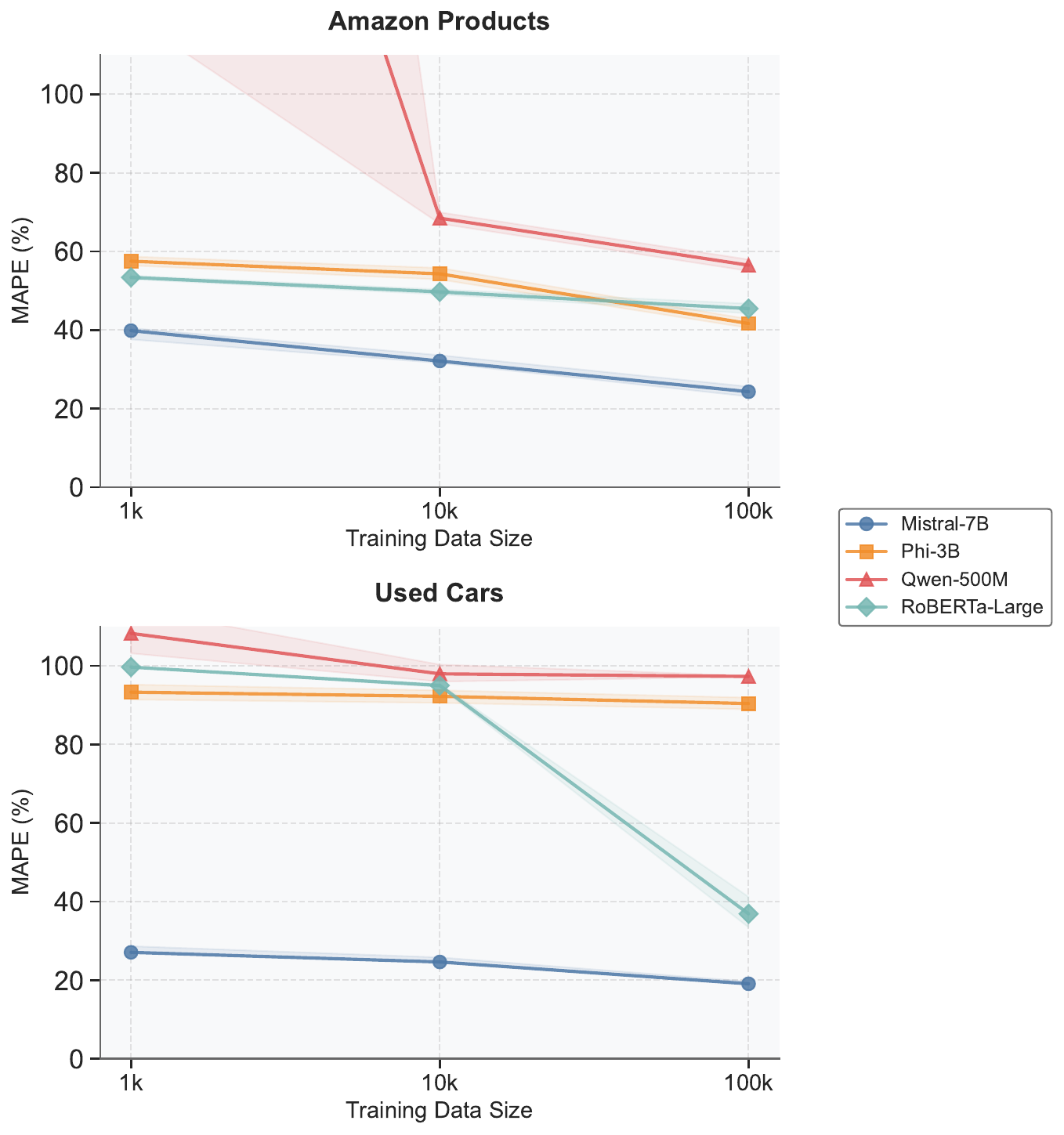}
    \caption{
    Impact of training data size on model MAPE on two datasets (y-axis scaled 0-100\% for comparison). %
    } %
    \label{fig:scaling_effects}
\end{figure}

\subsection{Discussion}

\paragraph{Theoretical Justification of Distributional Regression.} 

While \Cref{tab:model_comparison} shows consistent out-performance of distributional regression (Mistral-7B-Quantile) over point regression (Mistral-7B-Point) across all point metrics, we also theoretically discuss why multi-quantile LLM fine-tuning is superior to point-estimate fine-tuning in capturing uncertainty. Fine-tuning LLMs with Mean-Square Error (MSE) loss  trains them to learn the conditional mean, ignoring higher-order moments and distributional shape. This is because the gradient is proportional to the raw error $(\hat{y} - y)$, and all corrections push predictions towards the conditional mean. 
On the other hand, Pinball loss for a quantile $\tau$ yields a consistent estimator of that specific quantile \cite{koenker1978regression}. For each observation, when the model under-predicts the $\tau$-quantile, the gradient moves the prediction upward with weight $\tau$. Conversely, for over-prediction, the gradient pushes downward, with weight 1-$\tau$. In our multi-quantile approach, summing over multiple $\tau$ provides a discrete approximation to the integral of pinball losses over $\tau$ in (0,1). This integral corresponds to CRPS which is a strictly proper scoring rule for the entire distributions \cite{gneiting2007strictly}, so minimizing it recovers the ground truth conditional distribution.

From a multi-task learning perspective, our approach benefits from shared representation across quantile predictions; each quantile level effectively functions as a related but distinct prediction task. The quantile approach yields better point estimates through two mathematical mechanisms: first, the $\tau=0.5$ quantile loss's inherent robustness to outliers, and second, when training with multiple quantile levels simultaneously, the non-median quantile losses effectively serve as regularization terms for the median prediction task, constraining the model to perform well across the entire distribution.

\paragraph{Performance Breakdown by Category.}

Analysis of Mistral-7B's performance across different product categories in Amazon Products data reveals interesting patterns.%
The model excels in categories with standardized pricing structures, achieving MAPEs as low as 4.68\% for Window Tinting Kits which also has a high price range spanning \$200; and 5.39\% for Keyrings \& Keychains. 
While there are high performing categories with either narrow price ranges (e.g., Machine Screws: \$7.35-\$13.84) or well-defined market segments (e.g., Engine Management Systems), the model is also able to make good quality price predictions (MAPE within 6-12\%) for widely diverse categories with high price ranges (e.g., MAPE of 10.43\% for Custom Fit with a price range over \$500, and 12.11\% for Body with a price range over \$400).
More details can be found in Tables~\ref{tab:top_mape} and~\ref{tab:bottom_mape}. %

\paragraph{Performance Improvement with Training Data and Model Size.}
\Cref{fig:scaling_effects} illustrates clear performance gains %
with increased training data. %
Mistral-7B-Quantile shows strong scaling benefits for the Amazon Products dataset, with MAPE decreasing from 39.84\% at 1,000 samples to 24.3\% at 100,000 samples. The Used Cars dataset exhibits similar scaling behavior, with MAPE reducing from 27.09\% to 19.09\% across the same range.
Mistral-7B's superior performance over smaller models (Phi, Qwen and RoBERTa) across all metrics shows that model scale significantly impacts price estimation accuracy, particularly in complex scenarios.

\paragraph{Training Data Contamination.}
Multiple lines of evidence suggest our results may not be due to contamination between LLM pre-training and our test data.
State-of-the-art LLMs like Claude-3.5-Sonnet perform poorly (\Cref{tab:model_comparison}, Figure \ref{fig:few_shot_scaling}) without task-specific fine-tuning, indicating limited retention of price relationships during pre-training.
Additionally, our data scaling experiments (Figure~\ref{fig:scaling_effects}) show consistent performance improvements with increased training data across all datasets, indicating that more data contributes to higher learning.
However, we cannot provide definitive proof that our test data was not encountered during LLM pre-training, and contamination remains a possibility despite the evidence presented above.

\paragraph{Practical Applications of Text-To-Distribution Modeling.}

Our price distribution estimation approach from unstructured text input helps in generating substantially more informative outputs than traditional point-wise regression methods. These price distributions can be used for: (i) capturing varying degrees of price uncertainty across items, (ii) providing interpretable probability bounds (e.g., 90\% confidence intervals), and (iii) representing diverse distribution shapes as shown in \Cref{fig:pred_dist_plots} and \Cref{fig:pred_dist_plots_additional}.

\section{Conclusion}
\label{sec:conclusion}

We demonstrated the effectiveness of LLMs with quantile regression heads for probabilistic price prediction from unstructured inputs which not only produce well-calibrated price distributions but also achieve superior point estimates compared to traditional approaches. Our Mistral-7B-Quantile model outperforms traditional approaches and few-shot, in-context learning across multiple datasets, with performance notably improving with larger model sizes and training volumes. Our findings establish a foundation for probabilistic regression with LLMs and showcase their capability in complex numeric prediction tasks using unstructured input.

There are several promising avenues for future research, such as hybrid architectures combining decoder-only models with traditional pricing approaches, explicitly incorporating domain knowledge about pricing and market dynamics, exploring advanced LLM reasoning techniques, and building more interpretable and reliable models that provide insights into they make pricing decisions. 

The LLM-based regression approach could also be applied to a number of other existing text-based tasks, such as financial forecasting of return and volatility from news articles and social media, sentiment analysis, and text readability scoring.

\section{Limitations}

We acknowledge the following limitations of our study. First, we did not fine-tune LLMs larger than 7B parameters in size. Second, although we focused exclusively on the pricing task in this work, we believe that our quantile regression approach would generalize well to other domains, given that our model architecture contains no domain-specific components. However, we did not evaluate this on other general regression domains or non-price prediction tasks. Third, we do agree that our training data is old and it is possible that the LLMs we experimented upon may have seen this data during their pre-training. However, multiple experimental results discussed earlier suggest that this contamination does not significantly contribute to observed results. Finally, some of the listings in our datasets date back to 5-10 years, and we did not explore in detail how this can affect the performance of our in-context learning baselines. 

\section*{Acknowledgments}
The authors would like to thank Yiman Ren and Arman Akbarian for preliminary explorations of this project, and the anonymous reviewers for their helpful feedback. 

\bibliography{references}

\appendix
\onecolumn

\section*{\center Appendix}
\label{sec:appendix}

\section{Dataset Details and Examples}
We show in \Cref{tab:data_format_examples} examples of the inputs in each of our three datasets. Each data entry contains both structured and unstructured text information. The currency distribution of the Used Boats dataset is shown in Table \ref{tab:currency_dist}. We also show in Figure~\ref{fig:cleaningprompt} the LLM prompt that we used to clean up all three of our datasets and remove the rows containing erroneous (described in Section \ref{sec:data}). We show examples of such rows in Table \ref{tab:bad_prices}.  

\subsection{Validating LLM-based Data Filtering}\label{subsec:cleanup-eval}

To address potential concerns about whether the LLM filtering removed hard examples or created unintended biases, we made two key observations. First, we noted that Claude, the model performing the cleanup, shows poor performance in zero-shot and few-shot settings on the data it marked as clean, providing initial evidence that it did not selectively retain easily predictable cases.

More rigorously, our human evaluation study on a balanced subset comparing both LLM-accepted and LLM-rejected cases confirmed that the filtering criteria were appropriate and unbiased. Specifically,  we selected a balanced random subset of data marked as both acceptable and unacceptable by the LLM. Independent human evaluators assessed these samples without knowledge of the LLM's decisions. As shown in Table~\ref{tab:validation_results}, human evaluators assessed 341 samples from the Amazon dataset and 153 samples from the Cars dataset. The results demonstrate strong agreement between human and LLM judgments, with agreement rates of 95.3\% and 94.1\% for Amazon and Cars datasets, respectively.

To further validate the filtering effectiveness, we compared model performance on both LLM-filtered and human-validated subsets. For the Amazon dataset, Mistral-7B-Quantile achieved a MAPE of 16.3\% (95\% CI: [14.3\%, 18.3\%]) on LLM-filtered data and 43.76\% (95\% CI: [14.9\%, 87.5\%]) on human-validated data. For the Cars dataset, the model showed nearly identical performance with MAPE of 5.82\% (95\% CI: [4.23\%, 7.43\%]) and 5.79\% (95\% CI: [4.22\%, 7.53\%]) on LLM-filtered and human-validated sets, respectively.

A Fisher test with bootstrap sampling comparing the MAPEs between LLM-filtered and human-validated test sets yielded a p-value of $0.198$, indicating no statistically significant difference between the two sets' prediction accuracy. This statistical evidence, combined with the high human-LLM agreement rates and Claude's poor zero-shot performance on the filtered data, strongly supports the reliability and unbiased nature of our LLM-based filtering approach.

\begin{table*}
\small
\begin{tabular}{p{2.5cm}p{12cm}}
\toprule
\textbf{Dataset Type} & \textbf{Example Data Structure} \\
\midrule
\textbf{Amazon Products} & 
\begin{minipage}[t]{\linewidth}
\begin{verbatim}
{<product>
    <title>Tubing End Cap Solid Brass Scroll End</title>
    <description>CAP-off your railing in style with our selection 
    of END CAPS and PLUGS...</description>
    <brand>Renovator's Supply</brand>
    <type>Pipe Fittings</type>
    <attributes>Part Number: 95988, Material: Solid Brass</attributes>
</product>,
'price': $34.163}
\end{verbatim}
\end{minipage} \\
\midrule
\textbf{Used Cars} & 
\begin{minipage}[t]{\linewidth}
\begin{verbatim}
{<used_car>
    <model_type>pickup, sierra 1500 crew cab slt, gmc, 2014.0</model_type>
    <description>Carvana is the safer way to buy a car During these uncertain times,
    Carvana is dedicated to ensuring safety for all of our customers. In addition to 
    our ...[Removed due to length]</description><size></size><color>white</color>
    <region>auburn, , al</region><condition>good, clean</condition>
    <features>cylinders: 8 cylinders, fuel: gas, odometer: 57923.0,
    transmission: other, VIN: 3GTP1VEC4EG551563, drive: , </features>
</used_car>, 
'price': $33589.548}
\end{verbatim}
\end{minipage} \\
\midrule
\textbf{Used Boats} & 
\begin{minipage}[t]{\linewidth}
\begin{verbatim}
{<boat>
    <boat_type>Flybridge</boat_type>
    <boat_manufacturer>Galeon power boats</boat_manufacturer>
    <size>Length: 9.6, Width: 3.0</size>
    <condition>Used boat, Diesel</condition>
    <material>GRP</material>
    <region>Italy » Lombardia - Trentino Alto Adige » 
    MARINA DI VERBELLA - LAGO MAGGIORE</region>
    <year_built>2005</year_built>
    <price_currency>EUR</price_currency>
</boat>, 
'price': €68000}
\end{verbatim}
\end{minipage} \\
\bottomrule
\end{tabular}
\caption{Example data format for different datasets. Each dataset contains both unstructured and structured fields with categorical and numerical valued attributes, capturing various item attributes and price information.} %
\label{tab:data_format_examples}
\end{table*}

\begin{table}%
\centering
\small
\setlength{\tabcolsep}{4pt}
\begin{tabular}{lr}
\hline
Currency & Count \\
\hline
EUR & 8,430 \\
CHF & 980 \\
GBP & 298 \\
DKK & 180 \\
\hline
\end{tabular}
\caption{Used Boats Currency Distribution}
\label{tab:currency_dist}
\end{table}

\begin{table}%
\centering
\small
\setlength{\tabcolsep}{3pt}
\begin{tabular}{lrrrr}
\toprule
Dataset & Total & Both & LLM Acc., & LLM Rej., \\
 & Samples & Agree & Human Rej. & Human Acc. \\
\midrule
Amazon & 341 & 325 & 14 & 2 \\
Used Cars & 153 & 144 & 9 & 0 \\
\bottomrule
\end{tabular}
\caption{Human Validation of LLM-cleaned Prices. `Acc.' and `Rej.' stand for Accept and Reject.}
\label{tab:validation_results}
\end{table}

\begin{table*}%
\small
\caption{Examples of erroneous prices across datasets that were removed}
\label{tab:bad_prices}
\begin{tabular}{lp{0.15\textwidth}p{0.35\textwidth}p{0.1\textwidth}r}
\toprule
Dataset & Product Type & Description Summary & Condition & Price \\
\midrule
Amazon Products & RAM Memory & \makecell[l]{16GB (2x8GB) DDR3 RAM\\for Toshiba Satellite} & New & \$3.28 \\
Amazon Products & Window Insulation Kits & \makecell[l]{500 sqft (4ft x125ft) of NASA TECH\\Commercial Grade Reflective Insulation} & New & \$2.85 \\
Used Cars & Mercedes E-Class & 2015, 59,749 miles,4MATIC, Blue & Excellent & \$1.00 \\
Used Cars & Chevrolet Malibu LS Sedan & 2015, 79,539 miles, Blue & Clean & \$165 \\
Boats & Rigiflex Motor Yacht & 2017, 4m length, 1.9m width,Switzerland & New & 3337 CHF \\
Boats & Whaly Pontoon boat & 2018, 4.35m length, 1.73m width,Italy & New & 3300 EUR \\
\bottomrule
\end{tabular}
\end{table*}

\subsection{Price Distributions}
\begin{figure}%
\centering\includegraphics[width=0.9\textwidth]{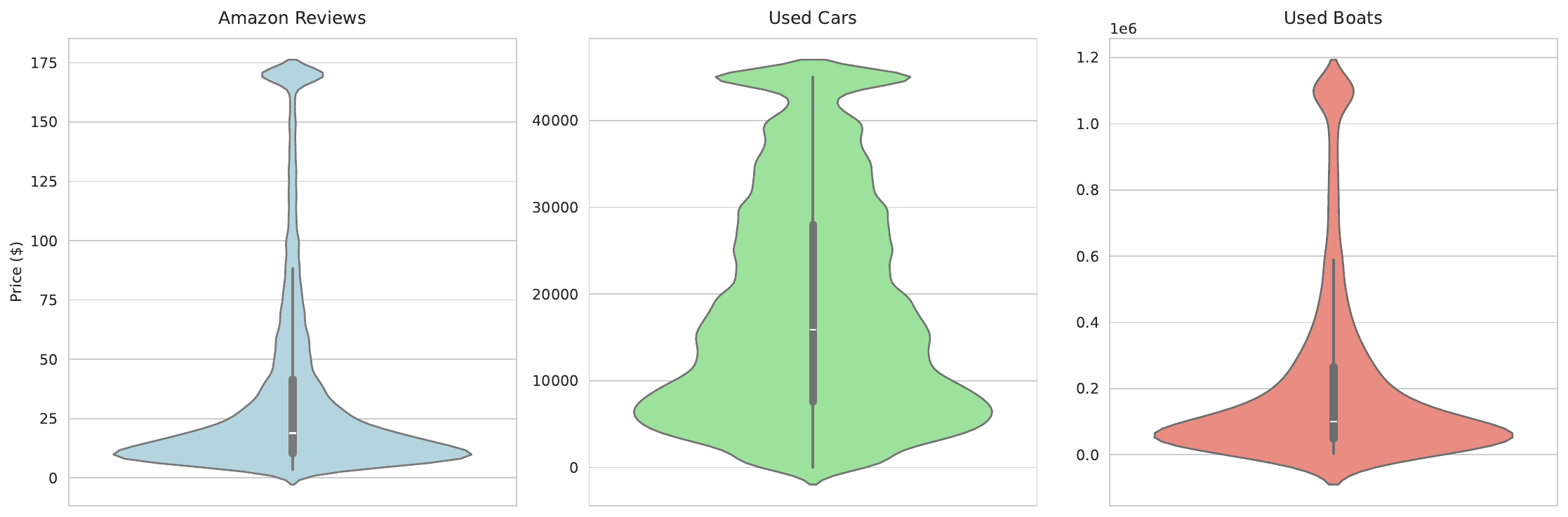}
    \caption{Density distribution of prices across three different datasets: Amazon Products, used cars, and used boats. The distributions are trimmed at the 95th percentile to handle outliers.}
    \label{fig:price_distributions}
\end{figure}

Figure~\ref{fig:price_distributions} presents the density distributions of prices across the datasets. All distributions exhibit notable right-skewed patterns, though with varying degrees of concentration and scale. The Amazon Products prices show a sharp peak around \$25 with a relatively narrow spread, suggesting most reviewed products fall within the affordable consumer goods range. The used car market displays a broader distribution centered approximately around \$15,000-\$20,000, with a gradual tapering toward higher price points. The used boat market demonstrates the largest price variation, with values extending into the millions of dollars, though the core distribution remains concentrated in the lower price ranges. In this visualization, all distributions are trimmed at the 95th percentile, to trim the outliers.

\section{Further Modeling Details}
\label{app:modelingdetails}

\subsection{Ensuring Monotonicity and Continuous Quantile Prediction}
\label{sec:quantile_monotonicity}

This section describes two structural additions we can implement in the quantile regression head, previously denoted by $g(\cdot;\phi)$ in \cref{subsec:quantile-head}, to ensure two properties. First, is the monotonicity of the quantile. Specifically, just a quantile regression head and the use of pinball loss provides no guarantee that predicted quantiles $\hat{q}_{\tau_1}, \hat{q}_{\tau_2}, \ldots, \hat{q}_{\tau_K}$ will satisfy the monotonicity constraint $\hat{q}_{\tau_i} \leq \hat{q}_{\tau_j}$ for $\tau_i < \tau_j$. This can lead to nonsensical predictions where, for example, the 90th percentile could be lower than the 80th percentile. The second issue is limited quantile resolution. That is, training on a fixed set of $K$ quantile levels (e.g., $\tau \in \{0.1, 0.2, \ldots, 0.9\}$) restricts predictions to these specific levels, preventing inference at arbitrary quantile levels such as $\tau = 0.73$.

Below, we describe how we can address both challenges through a combination of delta encoding and linear interpolation.

\paragraph{Monotonicity via Delta Encoding:} Instead of directly predicting quantile values, as in \cref{eq:output-of-llm} via a regression head, we can adjust the architecture to predict the first quantile value: $\hat{q}_{\tau_1}$, and the non-negative differences between consecutive quantiles: $\Delta_i = \hat{q}_{\tau_{i+1}} - \hat{q}_{\tau_i} \geq 0$. 

This can be implemented as:
\begin{align}
    \vz_{\text{deltas}} &= [\vz_0, \sigma(\vz_1), \sigma(\vz_2), \ldots, \sigma(\vz_{K-1})] \\
    \mathbf{\hat{q}} &= \cumsum(\vz_{\text{deltas}})\,,
\end{align}
where $\vz$ is $\vh_T$, $\sigma(\cdot)$ is a non-negative activation function (e.g., $\relu$ or $\softplus$), and $\cumsum$ denotes the cumulative sum operation. This construction guarantees $\hat{q}_{\tau_1} \leq \hat{q}_{\tau_2} \leq \ldots \leq \hat{q}_{\tau_K}$ by design.

Note that the above modification is purely an architectural modification that guarantees monotonicity by construction, while keeping the loss function and training objective exactly the same as described in \cref{sec:methods}. The network still learns to minimize the pinball loss, it just does so through an architecture that makes it impossible to violate monotonicity.

\paragraph{Continuous Quantile Prediction via Interpolation:} To predict quantiles at arbitrary levels $\tau \in (0,1)$ not in our initial quantile levels used during training, one could use linear interpolation between adjacent trained quantiles. For a query quantile $\tau$, one can find the adjacent trained quantile indices: $i = \lfloor \tau \cdot (K-1) \rfloor$ and $i + 1$, then compute the interpolation weight: $w = \tau \cdot (K-1) - i$, and interpolate: 
\[
\hat{q}_\tau = (1-w) \cdot \hat{q}_{\tau_i} + w \cdot \hat{q}_{\tau_{i+1}}\,.
\]
This leads to continuous quantile predictions across the entire range $(0,1)$ while maintaining monotonicity, as linear interpolation preserves order relationships.

\subsection{Few-shot Learning}

Few-shot learning enables models to make predictions with limited training examples, a capability that has proven particularly effective with LLMs \cite{wang2020generalizing}. Recent theoretical work has demonstrated that this ability, also known as in-context learning, has roots to transformer architectures \cite{garg2022what, bai2023transformers, vacareanu2024words}. 

In our pricing context, few-shot learning allows LLMs to leverage their pre-trained knowledge for price estimation with minimal additional examples. We enhance this approach by selecting prompt examples similar to the target product based on the category or manufacturer of the respective items, similar to retrieval-augmented generation (RAG) techniques \cite{lewis2021retrieval}. %

We evaluate the zero-shot and few-shot performance of two state-of-the-art LLMs, Claude-3.5-Sonnet and Nova Pro \cite{anthropic2024claude,aws2024nova}. We implement three few shot example selection strategies: (i) random sampling; (ii) category-based stratified sampling and (iii) similar item sampling based on cosine similarity of Qwen2-7B embeddings. The latter two 
leverage domain similarity for potentially better price estimation. We vary the number of examples as $\{0, 2^0, 2^2, \ldots, 2^{11}\}$, constrained only by the available dataset size and the LLM context window length, %
to analyze the relationship between example count and performance. All few-shot experiments use consistent prompts, shown in Figure~\ref{fig:fewshotprompt}, and temperature equal to $0$. Aligned with prior literature \citep{vacareanu2024from}, we utilize these models for point estimates only, as distributional predictions require specialized decoding rules \citep{lukasik2024regression} that are limited to open-source models.

\subsection{Using Cross-Entropy Loss}

In preliminary experiments, we compared three fine-tuning approaches: regression with squared error loss, regression with quantile (pinball) loss, and token prediction with cross-entropy loss. The regression approaches directly optimize for price predictions, treating the task as a continuous value prediction problem, while the cross-entropy approach treats prices as text and follows the traditional next token prediction.

In our experiments, regression-based approaches significantly outperformed the cross-entropy approach, with squared error loss showing a 1.11 percentage point improvement in MAPE (95\% CI: [0.40\%, 1.87\%]). Based on these findings, we focused on regression and quantile loss fine-tuning for all subsequent experiments.

\begin{table}[t!]
\footnotesize
\begin{center}
\setlength{\arrayrulewidth}{1.5pt}
\ttfamily
    \begin{tabular}{|p{0.95\linewidth}| }
    \hline

You are an expert in understanding product details and product prices. Given the below information about a product and its corresponding sale price, judge whether the given price is within a reasonable range for the given product, or if it is too high or too low. 

\\
Also generate a short reason. Your final output should be a single dict within <result> tags with two keys: price\_quality and reason.

\\

[PRODUCT INFO]
\\
Sale Price: [PRICE INFO]

\\ \hline
    \end{tabular}
    \vspace{1.0em}
    \captionof{figure}{Sample LLM prompt that we used to clean up the three of our datasets to remove rows with unreasonably high or unreasonably low prices, with respect to the item contexts.)}
    \label{fig:cleaningprompt}
\end{center}
\end{table}

\begin{table}[t!]
\footnotesize
\begin{center}
\setlength{\arrayrulewidth}{1.5pt}
\ttfamily
    \begin{tabular}{|p{0.95\linewidth}| }
    \hline
    You are an expert in understanding product details and product prices. \\

Predict the price in US dollars as a float32 number, for the given set of products. 
Output a JSON dict with a key for each input product ID, and a nested dict with a key 'price' containing your predicted price of the product, and another key 'reason' briefly explaining why your predicted price is correct.\\

Put the output JSON dict in <result> tags.\\
\\

Here are some examples of products and their prices. \\
\\

[EXAMPLES]

\\

Now predict the price for: \\

[CONTEXT]
\\ \hline
    \end{tabular}
    \vspace{1.0em}
    \captionof{figure}{Sample prompt for zero shot and few shot LLM based price prediction. This prompt is customized for the Amazon Products dataset, but we used very similar prompts for the other two datasets as well, with minor modifications (e.g., changing references to `products` to `used cars` etc.)}
    \label{fig:fewshotprompt}
\end{center}
\end{table}

\FloatBarrier

\section{Metric Definitions and Implementation Details}
\label{app:metrics_and_definitions}

We clarify that all publicly available datasets as well as models that we used in this work were used in accordance with their license and terms for use. We did not use any data or model outside of its intended purpose.

\paragraph{Quantile Levels and Point Prediction:} 
For all models that predict distributions, we take $K=200$ and $\vtau$ is obtained by dividing the interval $(0,1)$ into $K$ equal-length sub-intervals. We studied the impact of varying the number of quantiles $K = {10, 50, 200, 500, 1000}$ across the three datasets and found that initially as $K$ increases the performance improved, but plateaued after a certain point, which in our case was $K = 200$. We therefore used this number for all our experiments involving a trained model with a quantile regression head. 
We use models that produce a distribution both for generating probabilistic outputs and for point predictions. In the latter case, we take the predicted quantile at $\tau=0.5$ as the point estimate. Additionally, we include baseline models trained solely with traditional squared error loss, using their direct predictions for comparison. 

We also tune the value of the smoothing parameter $\alpha$, that controls how closely the $\softplus$ function approximates the $\relu$ function. We experimented with values ranging from $10^{-5}$ to $10^{-1}$ and did not observe significant effects. We therefore settled on $10^{-2}$, to achieve a balance between being closer to a true quantile loss and also achieving numerical gradient stability.

\subsection{Baselines}
\paragraph{Text Embedding Baselines:} 
We evaluate traditional ML models with text embeddings.
Text features (title, description, attributes) are concatenated with appropriate field markers and converted to embeddings using the general Qwen2-7B-instruct embedding model \citep{chu2024qwen2}. These embeddings serve as input features for five models: Ridge Regression and XGBoost for point estimation, Quantile Regression (with two hidden layers) for distribution prediction, trained on log-transformed target\footnote{In all three data sets since the target was price, we used its log-transformed prices to handle the wide range of values in our datasets during training.}, and two nearest neighbor-based distribution prediction approaches. The first nearest neighbor model predicts distributions by using the empirical distribution of target values from selected neighbors in the training set, while the second variant employs a radius-based selection criterion with a minimum neighbor requirement. All hyperparameters are selected using 5-fold cross-validation.

\paragraph{Fine-tuned Decoder LMs with Quantile Head:}
We fine-tune Mistral-7B (7 billion parameters), \cite{mistral2023mistral7b}, Phi-3B (3B parameters), \cite{abdin2024phi3}, and Qwen-500M (500M parameters), \cite{bai2023qwen}, using LoRA \cite{hu2022lora} (rank=192, alpha=384, dropout=0.1) with the AdamW optimizer \cite{loshchilov2018decoupled} (learning rate=$1.0\mathrm{e}{-06}$, weight decay=0.01), with the quantile head described in Section~\ref{sec:methods}, on log-transformed targets.

\paragraph{Fine-tuned Encoder LMs with Quantile Head:}
We fine-tune XLM-RoBERTa \cite{conneau2020unsupervised} in both base (279M parameters) and large (561M parameters) variants, adding a regression head as described in Section~\ref{sec:methods}. %

\paragraph{Fine-tuned LLM with Regression Head:}
We fine-tune Mistral-7B, the largest LLM in our set, with a regression head to study the impact of quantile prediction versus point estimation.

\paragraph{Few-shot SOTA LLMs:}
We evaluate the zero-shot and few-shot performance of two state-of-the-art LLMs, Claude-3.5-Sonnet and Nova Pro \cite{anthropic2024claude,aws2024nova}. For few-shot learning, we implement three example selection strategies: (i) random sampling; (ii) category-based stratified sampling and (iii) similar item sampling based on cosine similarity of Qwen2-7B embeddings. The latter two 
leverage domain similarity for potentially better price estimation. We vary the number of examples as $\{0, 2^0, 2^2, \ldots, 2^{11}\}$, constrained only by the available dataset size and the LLM context window length, %
to analyze the relationship between example count and performance. All few-shot experiments use consistent prompts (Figure~\ref{fig:fewshotprompt} of Appendix) and temperature equal to $0$. Aligned with prior literature \citep{vacareanu2024from}, we utilize these models for point estimates only, as distributional predictions require specialized decoding rules \citep{lukasik2024regression} that are limited to open-source models.

For the Amazon Products and Boats dataset, even with the best-performing category-based sampling strategy and optimal shot count (256), both Claude and Nova-pro achieve MAPEs more than 35\%, lagging significantly behind fine-tuned Mistral-7B's MAPE of 16.86\% and 21\% respectively.
The performance disparity is similarly stark in the Used Cars dataset. 
For the Used Cars dataset, while Mistral-7B achieves a MAPE of 6.3\%,  few-shot approaches struggle with much higher error rates: 
both Claude and Nova-pro show MAPEs between 230-245\% with random sampling and 290-305\% with category-based sampling. Nova-pro performs similarly poorly, with error rates consistently above 220\%.
For the Boats dataset, the gap narrows somewhat but remains substantial. Mistral-7B's MAPE of 21.2\% still outperforms the best few-shot results (Claude with random sampling at 35\% MAPE) by a considerable margin.
Choosing few shot examples similar to the target item based on pairwise cosine similarity using Qwen-7B-embeddings also gives a MAPE within 2-3\% of the random sampling strategy.

Our experiments also reveal an intriguing pattern in few-shot learning performance. 
Contrary to common intuition, our experiments also reveal that increasing the number of examples beyond a certain point %
starts degrading model performance. 
This finding challenges the conventional wisdom that more examples invariably lead to better few-shot performance. 
The degradation might be attributed to several factors, such as models' context window size limitations, potential interference between examples, or increased complexity in extracting relevant patterns from larger sets of examples. 
This non-monotonic behavior suggests that careful attention must be paid to the number and quality of examples used in price prediction tasks, and there exists an optimal window for few-shot learning, beyond which additional examples may interfere with the model's ability to effectively leverage the in-context information.
This observation has important implications for the practical application of few-shot learning in pricing tasks, suggesting that careful attention should be paid to the number of examples used rather than simply maximizing them.

\subsection{Evaluation Metrics}
\label{app:metrics}

We use two sets of metrics, one evaluating the estimated distributions generated by our quantile regression models and the other for point estimates. For each metric we report 95\% confidence intervals with bootstrap resampling (1000 iterations):
$\text{CI}_{95\%}(M) = [\hat{M}_{(0.025)}, \hat{M}_{(0.975)}]$
where $\hat{M}_{(q)}$ denotes the $q$-th quantile of the bootstrap distribution of metric $M$.

\subsubsection{Distribution Quality Metrics}

Assuming we have a test set of size $n$: ${(\vx_i, y_i)}_{i=1}^n$ and for each test point $\vx_i$, we have predicted quantiles,
$\hat{\vquant}_{\vtau}(\vx_i)=(\hquant_{\tau_1}(\vx_i) \le \dots \le \hquant_{\tau_K}(\vx_i))$.

\paragraph{Calibration Error (CE):}
CE measures how well predicted quantiles match their theoretical coverage:
$\mathrm{CE}
=
(1/K)\sum_{k=1}^{K}
|\widehat{\mathrm{coverage}}(\tau_k) - \tau_k|
$.
where $\widehat{\mathrm{coverage}}(\tau_k)$ is the empirical fraction of true values in the test set, below the $\tau_k$ quantile.

\paragraph{Continuous Ranked Probability Skill Score (CRPSS):}
This metrics is a scale-free version of the well-known CRPS which measures the integrated squared difference between predicted and true cumulative distribution functions:
\[
\mathrm{CRPS}=\frac{1}{n}\sum_{i=1}^n
\int_{-\infty}^{\infty}
\Bigl(\hat{F}_{\vx_i}(r)
- \mathbf{1}_{y_i \le r}\Bigr)^2 \, dr\,,
\]
where $\hat{F}_{\vx_i}$ is the estimated CDF using $\hat{\vquant}_{\vtau}(\vx_i)$. As a proper scoring rule, CRPS converges to zero if and only if the predicted distribution matches the true distribution \cite{gneiting2007strictly}. We report the (scale-free) skill score 
\[
\text{CRPSS} = 1-\left(\frac{\mathrm{CRPS}_\text{model}}{\mathrm{CRPS}_\text{reference}}\right)\,.
\]
where the reference is the empirical distribution of training targets.

\paragraph{Relative Confidence Interval Width (RCIW).}
RCIW measures the average width of predicted intervals relative to the true value:
\[
\text{RCIW}_\gamma = \frac{100}{n}\sum_{i=1}^n \frac{U_i^\gamma - L_i^\gamma}{|y_i|}
\]
where $[L_i^\gamma, U_i^\gamma]$ is the predicted $(1-\gamma)$ CI for $\vx_i$.
RCIW captures the sharpness of the distribution, where smaller values indicate a tighter interval.

\subsubsection{Point Estimate Metrics} We report:
MAPE (Mean Absolute Percentage Error) 
\[
(100/n)\sum_{i=1}^n \left|(y_i - \hy_i)/y_i\right|\,,
\]
WAPE (Weighted Absolute Percentage Error), 
\[
\frac{100 \sum_{i=1}^n |y_i - \hy_i|}{\sum_{i=1}^n |y_i|}\,,
\]
and
MPE (Mean Percentage Error): 
\[
(100/n)\sum_{i=1}^n (y_i - \hy_i)/y_i\,.
\]

\subsection{Computation Infrastructure}
We used the AWS EC2 infrastructure for running all our experiments. We estimate the use of about 2000 GPU hours for all our model training and evaluations. We also used an AI assistant to help with some parts of code writing.

\section{Detailed Analysis of Model Performance}\label{sec:detailed_analysis}

In this section, we provide an examination of our Mistral-7B-Quantile model's performance across different product categories and analyze the distributional patterns captured by the model. First, we present a breakdown of prediction accuracy by category, revealing which product types are most (least) challenging for price prediction. We then explore how the model captures various distributional shapes that reflect the underlying market dynamics of different products.

\subsection{Performance Breakdown by Category}

We provide detailed performance breakdown of our best model on the different categories of each dataset in Tables \ref{tab:top_mape} and \ref{tab:bottom_mape}.

\begin{table*}[t]
\centering
\small
\begin{tabular}{lrrr}
\toprule
Category & MAPE [\%] & Size & Price Range [\$] [Min, Median, Max] \\
\midrule
Camera Lenses & 34.75 [20.57, 44.61] & 6 & [7.78, 36.60, 294.95] \\
Tools & 33.92 [23.22, 44.19] & 6 & [7.99, 15.93, 84.95] \\
Bakeware Sets & 33.29 [25.07, 40.03] & 8 & [3.99, 8.09, 130.48] \\
Compressors & 33.09 [25.68, 39.96] & 13 & [29.99, 165.00, 395.65] \\
All-Purpose Labels & 30.19 [15.89, 43.41] & 7 & [4.99, 14.95, 33.29] \\
Platters & 29.86 [20.35, 38.93] & 6 & [14.99, 26.46, 69.99] \\
Pickups \& Pickup Covers & 29.86 [22.64, 36.16] & 9 & [6.04, 12.40, 219.00] \\
Lighting Assemblies & 29.66 [15.11, 44.26] & 6 & [14.98, 35.67, 173.43] \\
Hard Hat Accessories & 29.40 [17.66, 42.69] & 6 & [3.99, 4.99, 11.09] \\
Internal Hard Drives & 29.17 [19.69, 38.61] & 12 & [14.99, 52.50, 599.99] \\
\bottomrule
\end{tabular}
\caption{Examples of Categories with High Mistral-7B MAPE for Amazon Products dataset (Minimum Size > 5)}
\label{tab:top_mape}
\end{table*}

\begin{table*}[t]
\centering
\small
\begin{tabular}{lrrr}
\toprule
Category & MAPE [\%] & Size & Price Range [\$] [Min, Median, Max] \\
\midrule
Window Tinting Kits & 4.68 [2.74, 6.75] & 19 & [24.49, 39.49, 283.94] \\
Keyrings \& Keychains & 5.39 [2.59, 8.12] & 6 & [5.99, 8.09, 10.19] \\
CV Boots \& Joints & 5.69 [3.38, 8.31] & 11 & [11.50, 11.88, 69.99] \\
Exhaust & 5.80 [2.69, 9.22] & 8 & [15.72, 122.78, 719.48] \\
Machine Screws & 6.01 [2.65, 9.60] & 8 & [7.35, 9.96, 13.84] \\
Socket Wrenches & 6.30 [2.32, 11.21] & 6 & [8.51, 14.65, 108.38] \\
Engine Management Systems & 6.65 [3.72, 10.82] & 19 & [15.22, 69.95, 69.95] \\
Inkjet Printer Paper & 6.79 [3.66, 9.69] & 6 & [9.50, 29.42, 152.26] \\
License Plate Frames & 6.94 [2.62, 13.28] & 11 & [5.66, 16.99, 29.99] \\
Highball Glasses & 6.94 [3.08, 11.00] & 6 & [34.46, 47.27, 110.36] \\
Touchup Paint & 7.37 [5.76, 9.05]  & 68 & [8.25, 15.30, 71.92] \\
Keychains  &  9.70 [7.79, 11.83] & 71 & [5.79, 10.99, 55.99] \\
Frames & 9.91 [8.64, 11.18] & 231 & [4.99, 14.99, 95.00] \\
Custom Fit &  10.43 [9.19, 11.68] & 175 & [18.99, 119.00, 599.00] \\
Body  & 12.11 [10.63, 13.60]  & 136 & [6.99, 43.49, 409.85] \\
\bottomrule
\end{tabular}
\caption{Examples of Categories with Low Mistral-7B MAPE for Amazon Products dataset (Minimum Size > 5)}
\label{tab:bottom_mape}
\end{table*}

\subsection{Distributional Patterns in Price Predictions}\label{sec:addnl_dist_results}

\begin{figure*}[ht]
    \centering
    \includegraphics[width=.3\textwidth]{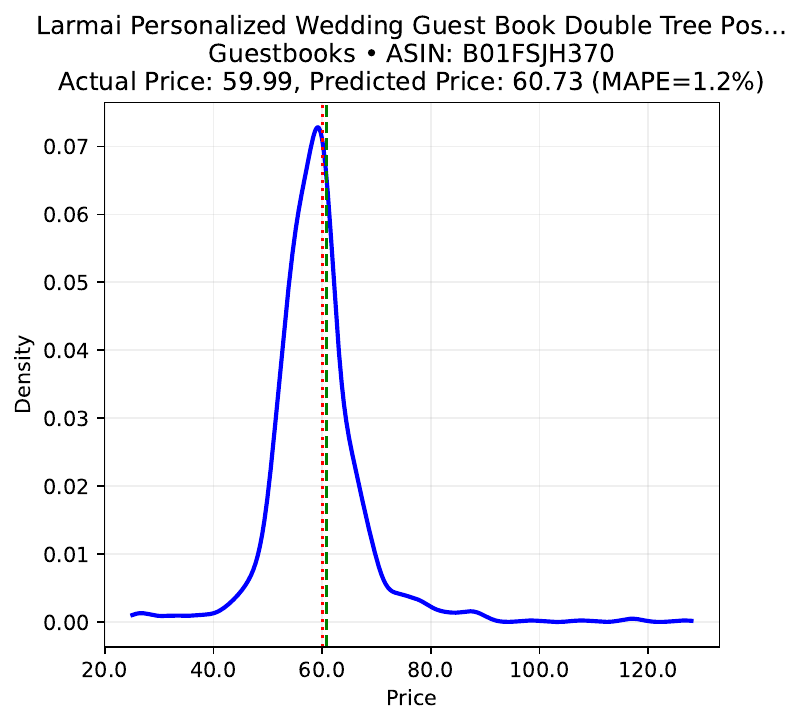}
    \includegraphics[width=.3\textwidth]{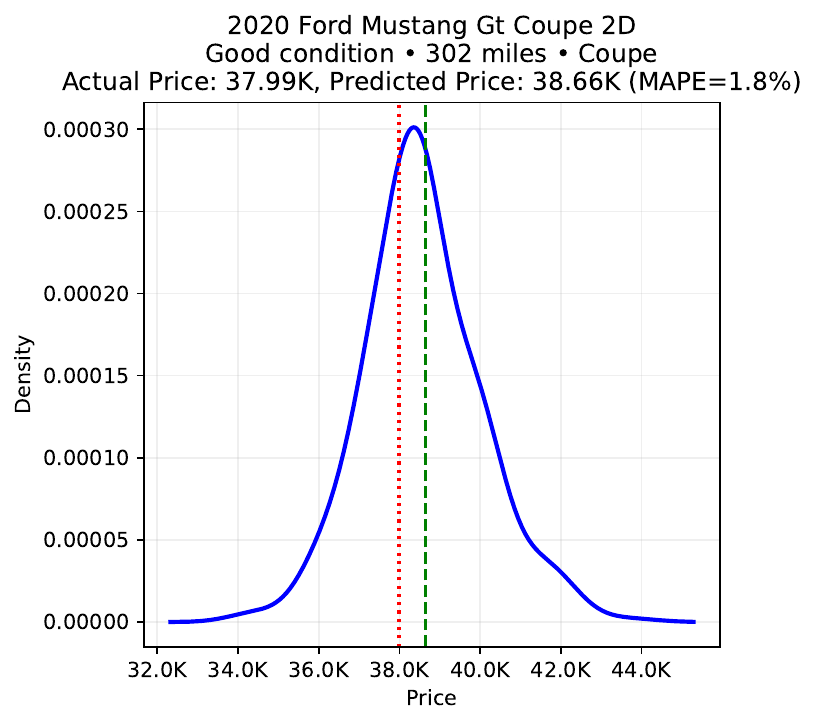}
    \includegraphics[width=.3\textwidth]{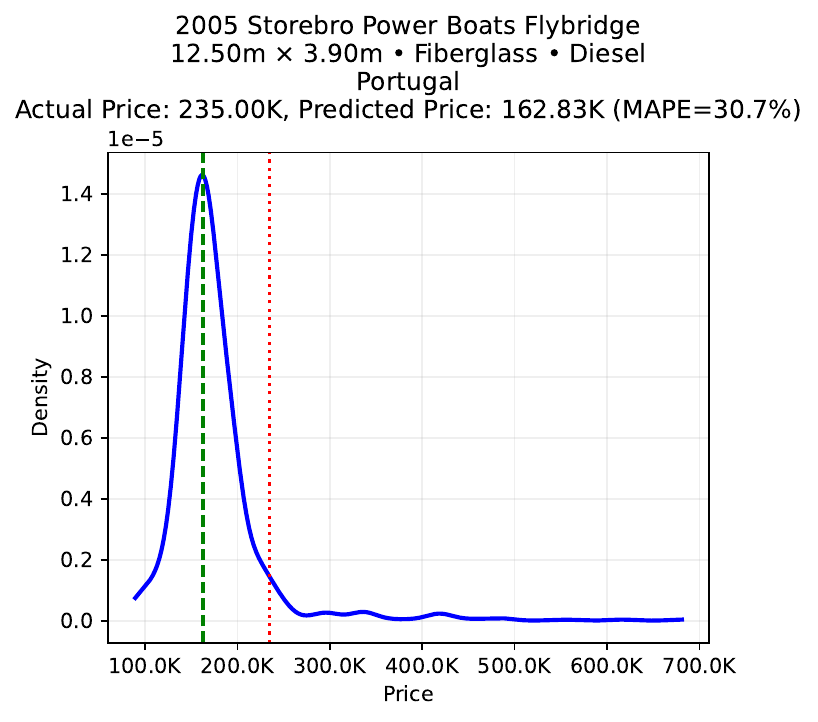}    
    \includegraphics[width=.3\textwidth]{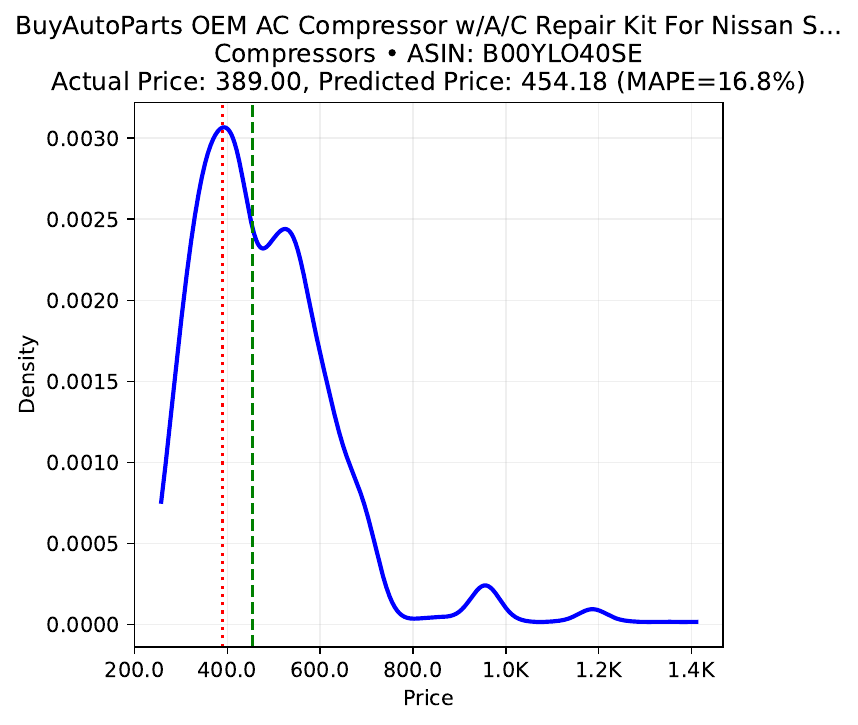}
    \includegraphics[width=.3\textwidth]{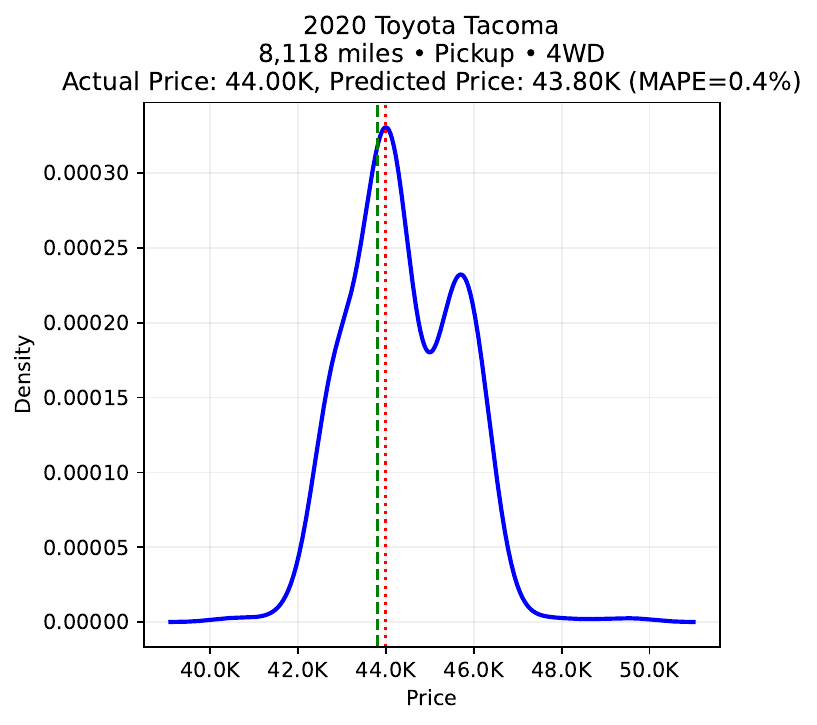}
    \includegraphics[width=.3\textwidth]{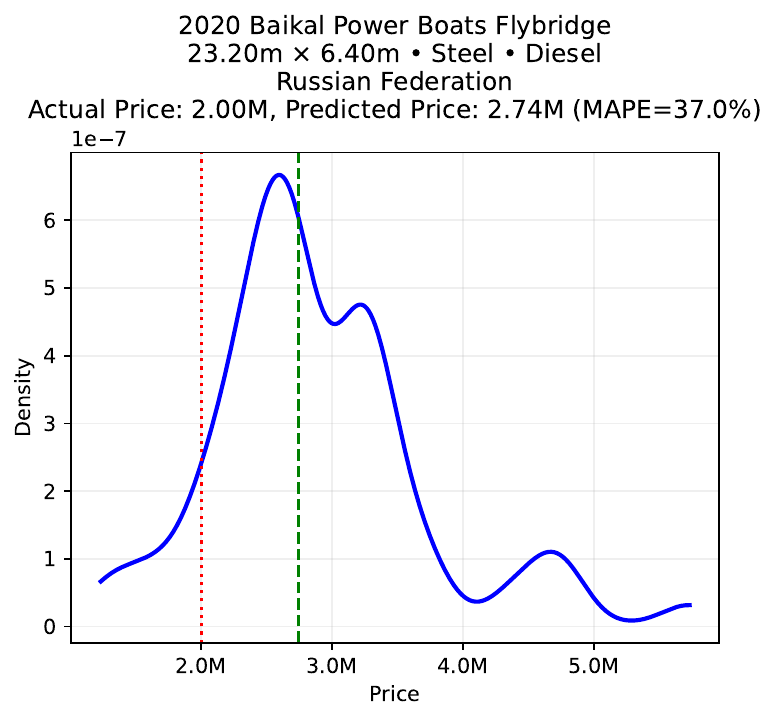}
    
    \caption{Probability density distribution of the prices predicted by the Mistral-7B-Quantile model across different datasets (\textcolor{blue}{blue} curve). Each x-axis has a different scale. The \textcolor{red}{red} dotted line represents the ground truth price while the \textcolor{mydarkgreen}{green} dashed line is the predicted median price. As demonstrated, the model captures different distribution shapes including unimodal (top row),  bimodal (bottom row), and right-skewed (right) distributions.}
    \label{fig:pred_dist_plots_additional}
\end{figure*}

The probability distributions shown in Figure~\ref{fig:pred_dist_plots} and Figure~\ref{fig:pred_dist_plots_additional} show different patterns that reflect the underlying market dynamics of different product categories.

\paragraph{Unimodal Distributions:} Products such as the Merritt tumbler, Toyota Corolla, Ford Mustang, and wedding guest book exhibit single-peaked distributions. These standardized products typically have well-established market prices with relatively low variance. The narrow, symmetric distributions suggest predictable pricing driven by clear market segments and standardized features, which is reflected in the model's lower prediction errors (MAPE ranging from 1.2\% to 6.5\%) for these items.

\paragraph{Bimodal Distributions:} Several products display dual peaks, including the Holley EFI gauges, AC compressor, and to varying degrees, the Lamborghini Huracán and Toyota Tacoma. This bimodality likely reflects distinct market segments. For automotive parts (gauges, compressor), the two peaks may represent new versus refurbished/used markets operating at different price points. For vehicles, different trim levels, model years, or condition categories (e.g., certified pre-owned versus standard used) create separate pricing clusters. Finally, the Toyota Tacoma's bimodal pattern potentially captures the price gap between base work trucks and fully-loaded consumer models.

\paragraph{Right-Skewed Distributions:} The luxury marine vessels (Sunseeker yacht, Storebro, and Baikai flybridge) exhibit right skew with heavier tails. This pattern aligns with the characteristics of high-end markets where, base models establish the primary peak, extensive customization options, rare features, or pristine/collector conditions create the long tail.
Additionally, the extreme tail (particularly visible in the Baikai boat with prices reaching \$500K+) likely represents highly customized or rare configurations.

The correlation between distribution shape and prediction accuracy is noteworthy. Standardized products with unimodal distributions achieve lower prediction errors, while luxury items with complex, skewed distributions show higher uncertainty (MAPE up to 37\%), that could be due to the inherent difficulty in pricing highly variable, customized products.

\end{document}